\documentclass[10pt,twocolumn,letterpaper]{article}

\usepackage[pagenumbers]{cvpr} 
\usepackage[linesnumbered,titlenumbered,ruled,vlined,commentsnumbered,noend]{algorithm2e}
\usepackage[table,xcdraw]{xcolor}
\usepackage{xcolor}
\usepackage{amsmath}
\usepackage{array} 
\usepackage{multirow} 
\usepackage{colortbl}
\usepackage{float}

\usepackage{placeins}
\usepackage{algorithmicx}
\usepackage{algpseudocode}
\definecolor{darkgreen}{rgb}{0.0, 0.5, 0.0} 
\newcommand{\red}[1]{{\color{red}#1}}
\newcommand{\green}[1]{\textbf{\color{darkgreen}#1}}

\definecolor{cvprblue}{rgb}{0.21,0.49,0.74}
\usepackage[pagebackref,breaklinks,colorlinks,allcolors=cvprblue]{hyperref}

\title{Anti-I2V: Safeguarding your photos from malicious image-to-video generation}

\author{
Duc Vu \quad Anh Nguyen \quad Chi Tran \quad Anh Tran \\
\small{Qualcomm AI Research$^{\dagger}$
}\\
\texttt{\scriptsize \{ducvu, anng, chitran, anhtra\}@qti.qualcomm.com} \\
}

\begin{document}
\maketitle
\newcommand\blfootnote[1]{%
  \begingroup
  \renewcommand\thefootnote{}\footnote{#1}%
  \addtocounter{footnote}{-1}%
  \endgroup
}

\makeatletter
\def\blfootnote{\gdef\@thefnmark{}\@footnotetext}
\makeatother
    
 \blfootnote{%
  \hspace{-1.7em}$\dagger$ Qualcomm AI Research is an initiative of Qualcomm Technologies, Inc.%
}

\begin{abstract}
Advances in diffusion-based video generation models, while significantly improving human animation, poses threats of misuse through the creation of fake videos from a specific person's photo and text prompts. Recent efforts have focused on adversarial attacks that introduce crafted perturbations to protect images from diffusion-based models. However, most existing approaches target image generation, while relatively few explicitly address image-to-video diffusion models (VDMs), and most primarily focus on UNet-based architectures. Hence, their effectiveness against Diffusion Transformer (DiT) models remains largely under-explored, as these models demonstrate improved feature retention, and stronger temporal consistency due to larger capacity and advanced attention mechanisms. In this work, we introduce Anti-I2V, a novel defense against malicious human image-to-video generation, applicable across diverse diffusion backbones. Instead of restricting noise updates to the RGB space, Anti-I2V operates in both the $L$*$a$*$b$* and frequency domains, improving robustness and concentrating on salient pixels. We then identify the network layers that capture the most distinct semantic features during the denoising process to design appropriate training objectives that maximize degradation of temporal coherence and generation fidelity. Through extensive validation, Anti-I2V demonstrates state-of-the-art defense performance against diverse video diffusion models, offering an effective solution to the problem.
\end{abstract}
\section{Introduction}
\label{sec:intro}
Recent advances in Video Diffusion Models (VDMs) \cite{sora, svd, gen3, modelscope, vcrafter1, vcrafter2, cog, animatediff} enable realistic and coherent video generation from text prompts, and many models \cite{cogX, kling, i2vgen, gen3, animatelcm} further support image-based animation that preserves fine visual details while generating motion aligned with text. Despite impressive capabilities, these models pose significant misuse risks, as a single reference image can be used to create deepfakes or harmful content.

Adversarial attacks have been explored as a defense by subtly perturbing input images to disrupt diffusion models. However, existing methods primarily target text-to-image or image-to-image generation \cite{antidb, advdm, mist, glaze, duaw, sds}, typically by maximizing denoising loss or degrading Variational AutoEncoder (VAE) \cite{vae} features. These approaches are insufficient for image-to-video generation, where the input image serves as the first frame and the core challenge is disrupting temporal consistency. Moreover, video diffusion models are larger and more resistant to adversarial noise.

In contrast, adversarial attacks for image-to-video generation remain underexplored. VGMShield \cite{vgm} perturbs reference images by disrupting image and video encoder features in Stable Video Diffusion (SVD), but merely distorting embeddings fails to effectively corrupt fine-grained details (e.g., human features) and generalizes poorly beyond SVD. DORMANT \cite{dormant} targets pose-driven human animation by disrupting appearance features using CLIP and ReferenceNet while enforcing frame incoherence loss; however, it requires pose guidance, limiting its applicability to pose-driven models \cite{animateanyone, magicanimate}. More recent methods, such as Vid-Freeze \cite{vidfreeze} and I2VGuard \cite{i2vguard}, suppress attention outputs through optimization-based objectives, but require high-end GPUs for effective optimization.

Although existing methods show strong performance on specific models, their effectiveness on recent large-scale image-to-video frameworks remains unclear. State-of-the-art models such as CogVideo-X \cite{cogX} and Open-Sora \cite{opensora} adopt Diffusion Transformer (DiT) or Multi-Modal DiT (MM-DiT) architectures with larger capacity and stronger temporal modeling, leading to more robust and higher-quality video generation. While Vid-Freeze \cite{vidfreeze} and I2VGuard \cite{i2vguard} evaluate CogVideo-X, conventional DiT-based frameworks like Open-Sora remain underexplored. Hence, we identify two key limitations in current attacks: \textbf{(1) Perturbation optimization space:} Most existing methods optimize perturbations in RGB space, primarily altering pixel intensities rather than deeper representations, making them easy to remove during denoising. Image transformations and purification methods \cite{diffpure, gridpure, impress} can further suppress such noise while preserving semantic content. \textbf{(2) Propagation of features through layers:} Prior approaches \cite{mist, advdm} typically target only the final outputs of components such as the text encoder, VAE, or denoiser, overlooking how features propagate across layers within the VAE and UNet or Transformer-based denoising modules.

To address these challenges, we propose \textbf{Anti-I2V}, a defense framework designed to prevent unauthorized image usage in diverse image-to-video diffusion models. First, we explore the impact of updating perturbations in the $L$*$a$*$b$* color space, focusing on the decorrelated $a$* and $b$* channels, in combination with the frequency domain. This approach offers an effective alternative to the conventional reliance on spatial RGB space. Second, Anti-I2V disrupts semantic feature extraction and propagation within the denoising network, explicitly at intermediate layers containing high-level representations. Furthermore, unlike previous work \cite{mist, advdm} focusing solely on the final encodings, we introduce layer-wise semantic disruption across both the VAE and denoising modules. Lastly, we evaluate Anti-I2V on a human-focused image-to-video benchmark, using various state-of-the-art video generation backbones, confirming the effectiveness and versatility of the proposed algorithm. Our main contributions are summarized below:
\begin{itemize}
    \item We introduce \textbf{Anti-I2V}, a method to effectively prevent unauthorized image usage in image-to-video generation across multiple varieties of diffusion models.
    \item We are the \textbf{first} to update perturbations in both the \textbf{\boldmath{$L^*a^*b^*$} color space} and \textbf{frequency domain}, demonstrating enhanced effectiveness and robustness.
    \item We analyze the semantic representations at individual layers within the denoising and VAE modules and introduce two tailored objectives, \textbf{Inter Representation Collapse (IRC)} and \textbf{Inter Representation Anchor (IRA)}. These objectives disrupt the internal feature representations of the diffusion model throughout the denoising process, leading to effective degradation of temporal coherence and generation fidelity.
\end{itemize}
\section{Related Works}
\label{sec:related}
\subsection{Video Diffusion Model}
Diffusion models (DMs) \cite{ddpm, ldm} have enhanced the quality and realism of image generation \cite{sdxl, flux, sb, sbv2, snoopi, sc}. In recent years, efforts have been made to extend these advancements to video diffusion models (VDMs), enabling the generation of high-fidelity and temporally coherent videos. Early UNet-based diffusion models operated directly in the pixel space by introducing a space-time factorization approach, which decouples intra-frame spatial feature extraction from inter-frame temporal modeling \cite{vdm}. To improve computational efficiency, later research shifted towards latent space operations or hybrid pixel-latent methods \cite{show1}. Many of these initial text-to-video (T2V) models enhanced large, pre-trained text-to-image (T2I) models by adding (2+1)D attention layers to manage spatial and temporal information \cite{align, modelscope, vcrafter1, vcrafter2, i2vgen}. More recently, the introduction of Diffusion Transformers (DiT) and Multimodal Diffusion Transformers (MMDiT) leads to state-of-the-art models like CogVideoX and OpenSora \cite{cogX, opensora}. These larger, highly scalable models utilize 3D full or sparse attention and temporally aware autoencoders, resulting in significantly robust and improved video generation. Alongside text-to-video, image-to-video generation has emerged as a key area, using images as conditional guidance to achieve higher-quality results with more precise control over characters and backgrounds \cite{dynamic, cogX, i2vgen, opensora, soraplan, ltx}. While this has led to realistic animations, it also presents a serious threat of misuse. The ability to generate unauthorized or harmful videos of individuals from photos sourced from the Internet or public datasets \cite{celebvtext,efhq} underscores the urgent need for effective protective mechanisms against misuse.

\subsection{Adversarial Attacks}
\textbf{Image Cloaking.} Image cloaking is an adversarial defense technique that introduces imperceptible perturbations to an input image $x$, producing a protected version $x_{adv}$ that disrupts target model behavior. Initially developed for face recognition systems \cite{fawkes}, it has recently been extended to diffusion-based image and video generation.

\noindent\textbf{Defenses for Image Generation.} Early works, such as AdvDM \cite{advdm} and Anti-DreamBooth \cite{antidb}, apply cloaking to prevent misuse in personalized text-to-image diffusion models. AdvDM targets Textual Inversion \cite{textualinv} by generating perturbations that interfere with textual-space personalization, while Anti-DreamBooth focuses on finetuning-based DreamBooth \cite{dreambooth} through Alternating Surrogate and Perturbation Learning (ASPL). Building on these works, Mist \cite{mist} introduces complementary semantic and textural losses to disrupt the diffusion process and VAE encoder. DiffProtect \cite{sds} improves efficiency by leveraging Score Distillation Sampling \cite{sds3d} to expose encoder vulnerabilities while reducing computational cost, while SimAC \cite{simac} incorporates selective denoising timesteps during optimization. Additionally, several approaches \cite{erase,suma,nguyen2025cgce} rely on model finetuning, which is beyond the scope of our study.

\noindent\textbf{Defense for Video Generation.}
To our knowledge, few cloaking techniques specifically target image-to-video generation. VGMShield \cite{vgm} perturbs reference images to disrupt feature representations within the image and video encoders of Stable Video Diffusion (SVD). However, its simple embedding distortions are often insufficient for concealing human features and perform poorly on non-SVD models. DORMANT \cite{dormant}, in contrast, targets pose-driven human image animation by disrupting appearance feature extraction using CLIP and ReferenceNet, and introduces a frame incoherence loss. Its reliance on reference pose images and CLIP-based components limits applicability to pose-driven or face-animation models, while its multi-model optimization demands high-end GPUs (e.g., A800 80GB), making it impractical for standard hardware. Vid-Freeze \cite{vidfreeze} designs losses that suppress either cross-attention weights or all attention weights, while I2VGuard \cite{i2vguard} introduces an adversarial loss on self-attention outputs to reduce temporal inconsistency. However, these attention-based objectives require substantial computational resources, demanding high-end GPUs (60–80 GB of memory) per optimization.

\noindent\textbf{Perturbation Optimization Space.} Most image cloaking methods modify perturbations directly in the RGB color space. Recent works improve attack effectiveness by transforming images into alternative domains to identify more influential pixels. InMark \cite{inmark} uses the Discrete Cosine Transform (DCT) \cite{dct} to locate key pixels in the low-frequency subspace, while HF-ADB \cite{hfdb} applies a high-pass filter to inject noise into high-frequency regions. Anti-Forgery \cite{antiforgery} operates in the $L^*a^*b^*$ color space, perturbing the decorrelated $a^*$ and $b^*$ channels to disrupt models. However, few studies have explored combining multiple non-RGB spaces for joint perturbation optimization.

\section{Preliminaries}
\textbf{Diffusion Model} is a probabilistic generative model that generate samples by drawing from a Gaussian distribution and progressively denoising them to approximate the target data distribution. Starting from an initial point $x_0$ drawn from a target distribution $q_0(x_0)$, the forward process gradually adds noise at each timestep $t \in [0, T)$, resulting in a sequence of noisy samples $\{x_0, x_1, \dots, x_T\}$. At $t=T$, $x_T$ follows an isotropic Gaussian distribution, $x_T \sim \mathcal{N}(0, I)$. The reverse process iteratively removes noise estimated using a trained network \( \epsilon_{\theta}(x_t, t, y) \), where \(y\) is the conditioning input, e.g., text, image, or both. The network is trained to predict the noise $\epsilon \sim \mathcal{N}(0,1)$ added in the forward process, by minimizing the following objective:
\begin{equation}
    \mathcal{L}_{DM}(x_0) = \mathbb{E}_{x_0,y, t \sim \mathcal{U}(0,T), \epsilon \sim \mathcal{N}(0,1)} \|\epsilon_{\theta}(x_t, t, y) - \epsilon\|_2^2,
\end{equation}

\noindent\textbf{Adversarial Attacks} generate a subtly perturbed image $x'$ that is visually indistinguishable from the original image $x \in \mathbb{R}^{C\times H \times W}$ but causes a target model $f$ to misclassify it. Such attacks are either untargeted, where $f(x') \neq y_{\text{true}}$, or targeted, where $f(x') = y_{\text{target}} \neq y_{\text{true}}$.

Within the context of diffusion models, these attacks optimize an adversarial perturbation $\delta_{\text{adv}}$, constrained by a maximum magnitude $\eta$ to maintain visual similarity, to disrupt the diffusion process:
\vspace{-2mm}
\begin{equation}\label{attack}           
\delta_{\text{adv}} = \arg\mathop{\max}\limits_{\left\Vert \delta \right\Vert_{p} < \eta} \mathcal{L}_{DM}(x+\delta),
\vspace{-2mm}
\end{equation}
where $\mathcal{L}_{DM}$ represents the diffusion loss function. A widely adopted optimization technique is Projected Gradient Descent (PGD) \cite{pgd}, iteratively refining the adversarial perturbation, as shown below:
\vspace{-2mm}
\begin{equation}\label{pgd}  
\begin{split}
x^{t+1} &=\Pi_{(x, \eta)}\left(x^t+\alpha \operatorname{sgn}\left(\nabla_x \mathcal{L}_{DM}(x+\delta)\right)\right),
\end{split}
\vspace{-2mm}
\end{equation}
where $x^0 = x$, $\alpha$ is the step size, $t$ is the iteration, $\operatorname{sgn}(\nabla_x \mathcal{L})$ is the gradient sign, and $\Pi_{(x, \eta)}$ projects onto the $\eta$-ball around $x$ for imperceptibility.

\noindent\textbf{CIELAB color space ($L^*a^*b^*$)} is a perceptually uniform color representation with 3 channels: $L^*$ for luminance ($0$–$100$), $a^*$ for green–red ($-128$–$127$), and $b^*$ for blue–yellow ($-128$–$127$). Conversion from RGB assumes D65 illuminant and requires prior non-linear sRGB linearization. Perturbations in $a^*$ and $b^*$ modify color perception without affecting brightness, making them less perceptible to humans yet effective in misleading models.

\noindent\textbf{Discrete Cosine Transform (DCT)} converts image data from the spatial domain to the frequency domain, represented as a sum of sinusoidal components with different frequencies and amplitudes. DCT compacts most visually important information into a few coefficients, making it effective for identifying influential pixels.

\begin{figure*}[ht]
    \centering
    \includegraphics[width=\linewidth]{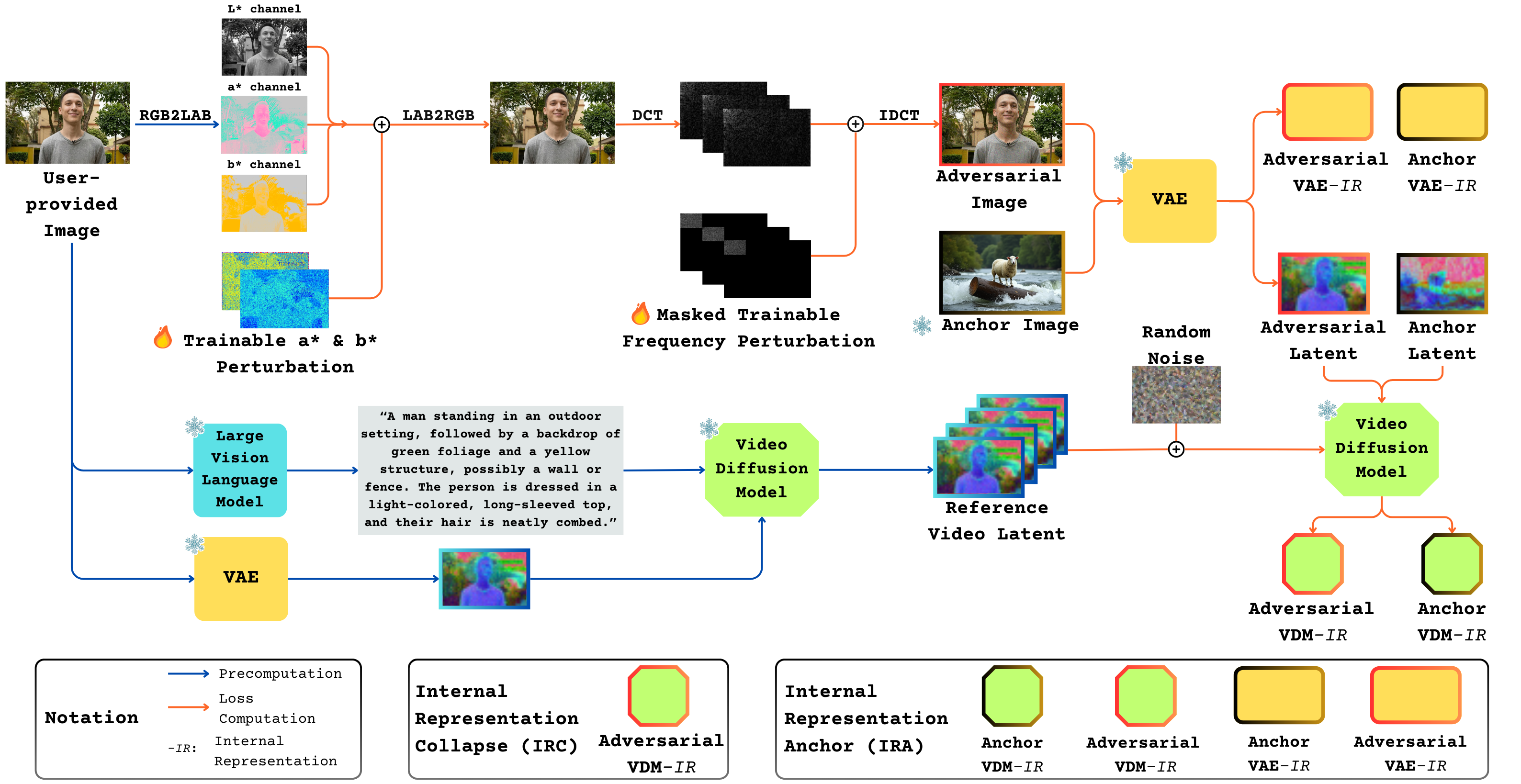}
    \captionsetup{aboveskip=3pt}  
    \caption{\textbf{Overall pipeline of Anti-I2V}. We first generate a reference video input by captions from leverage LVLM \cite{chen2024sharegpt4video} and user-provided image. Then we integrate the IRA and IRC losses with the vanilla training loss as the final objective. The noise is iteratively optimized through both $L^*a^*b^*$ and frequency space.}
    \vspace{-10pt}
    \label{fig:system}
\end{figure*}

\section{Method} 
\subsection{Problem Definition}
Given a reference image $x$ and a text prompt $y$, text-image to video diffusion models can be misused to generate harmful content. To prevent this, we craft an imperceptible perturbation $\delta$ for the reference image to degrade the model's generative capability. For any prompt $y$, videos generated from $x_{\xi}=x+\delta$ are distorted or semantically misaligned. Let $\epsilon_\theta$ denote the target video diffusion model, we aim to find $\delta$ by solving the following optimization problem:

\begin{equation}
    \begin{aligned}
        \delta^*= \arg\min_{\delta} \mathcal{L}_{Anti-I2V}(\epsilon_\theta, x + \delta, y) \\
        \text{s.t.} \|\delta\|_p \leq \Delta_{RGB},
    \end{aligned}
\end{equation}
where $\Delta_{RGB}$ denotes the adversarial noise budget for $\delta$. Optimizing $\delta$ is challenging because video models enforce strong semantic and structural coherence across frames through attention mechanisms. In image-to-video generation, the reference image forms the first, perceptually highest-quality frame. Unlike attacks on text-to-image models, our goal is to disrupt this temporal coherence so that subsequent frames are degraded.

\subsection{Overall proposal}
\label{subsec:proposal}
Our protective method combines two core components: a robust dual-space perturbation strategy (DSP) and the objective $\mathcal{L}_{Anti-I2V}$, as illustrated in \cref{fig:system}. Unlike image-generation tasks, image-to-video settings require a corresponding latent input of images and video. To address this, we generate a reference video from an image caption synthesized by an LVLM~\cite{chen2024sharegpt4video} using the target diffusion model $\epsilon$. Perturbation updates for $\delta$ follow the Dual-Space Perturbation method in \cref{subsec:dsp}, while the additional training objectives are detailed from \cref{subset:IRC} to \cref{subsec:final}.

\noindent\textbf{Robust Perturbation Space.} Most protection methods generate perturbations in the RGB space. However, the choice of perturbation optimization space has been largely underexplored. While RGB-space optimization is common, naïve perturbations applied directly in this space are  neutralized by video diffusion models. These models perform iterative multi-step denoising, progressively refining latent representations, and enforce strong spatio-tem nporal coherence through attention mechanisms that effectively suppress pixel-level perturbations. Therefore, an effective perturbation must target deeper representational levels beyond the pixel space. To this end, we introduce \textbf{Dual-Space Perturbation (DSP)}, a strategy designed to produce robust yet visually imperceptible adversarial perturbations.

\noindent\textbf{Temporal Coherence Disruption.} In image-to-video models, temporal attention mechanisms propagate information across frames, making the quality of each frame critically dependent on the semantic features extracted from preceding ones. Consequently, high-fidelity intermediate representations are essential for maintaining temporal coherence. We exploit this dependency by introducing targeted feature degradation to disrupt temporal consistency. To this end, our objective function, $\boldsymbol{\mathcal{L}_{Anti-I2V}}$, applies carefully crafted perturbations to corrupt feature representations at semantically rich layers. This disruption cascades through the attention mechanism, ultimately degrading both semantic and visual continuity in the generated video.

\subsection{Dual-Space Perturbation}
\label{subsec:dsp}
To improve perturbation robustness, we go beyond the standard RGB space, which has been shown to be insufficient \cite{antiforgery,inmark}. Following \cite{antiforgery}, we operate in the $L^*a^*b^*$ color space, which better aligns with human perception. Specifically, we convert the image to $L^*a^*b^*$ space, apply adversarial noise only to the $a^*$ and $b^*$ channels, before converting back to RGB. This approach yields less perceptible perturbations, strengthening the defense under a fixed budget while improving robustness against common transformations like blurring and JPEG compression.

Apart from $L^*a^*b^*$ space, another direction is to change RGB pixels based on signals from the frequency domain \cite{hfdb}. Inspired by \cite{inmark}, we focus on influential low-frequency components by analyzing the top-left coefficients of the Discrete Cosine Transform (DCT) \cite{dct}, since most content-irrelevant information resides in high-frequency signals. However, unlike \cite{inmark}, which uses frequency analysis to guide RGB-domain updates, we directly introduce adversarial noise in the frequency domain by perturbing the top-left DCT coefficients corresponding to the most influential low-frequency components.

This frequency-domain approach yields perturbations that are simultaneously more targeted in disrupting feature propagation and less perceptible spatially. By directly altering the core structural and textural information encoded in low-frequency components, the resulting perturbations are more effective and resilient, as they disrupt the fundamental representations of the image rather than the superficial values of pixels. Our final Dual-Space Perturbation (DSP) method integrates two stages: \textbf{(1)} Updates in the $L^*a^*b^*$ color space and \textbf{(2)} Updates in the DCT low-frequency domain. Full algorithm is detailed in \cref{alg:procedure}.

\begin{algorithm}[t]
    \small
    \SetAlgoLined
    \SetKwInput{Input}{Input}
    \SetKwInput{Output}{Output}

    \Input{Input image $x$, text prompt $y$, total noise budget $\Delta_{RGB}$, $L^*a^*b^*$-space budget $\Delta_{lab}$, frequency mask $M \in \{0,1\}^{H \times W}$, step size $\alpha$, diffusion model $\epsilon_\theta$, number of iterations $N$, and objective function $\mathcal{L}_{\text{Anti-I2V}}$}
    \Output{Adversarial sample $x_{\xi}$}

    \textbf{Initialize:}
    $\delta_{a^*}, \delta_{b^*} \in \mathbb{R}^{H \times W} \gets \mathbf{0}$\;
    $\delta_{\mathrm{freq}} \in \mathbb{R}^{C \times H \times W} \gets \mathbf{0}$\;

    \For{$i \in \{1,\dots,N\}$}{
        \textbf{$\triangleright$ Convert to $L^*a^*b^*$ space}\;
        $l^*, a^*, b^* \leftarrow rgb2lab(x)$\;

        \textbf{$\triangleright$ Add $L^*a^*b^*$-space perturbations to channels $a^*$ and $b^*$}\;
        $\delta_{a^*} \leftarrow clip(\delta_{a^*}, -\Delta_{lab}, \Delta_{lab})$\;
        $\delta_{b^*} \leftarrow clip(\delta_{b^*}, -\Delta_{lab}, \Delta_{lab})$\;
        ${a^*}' \leftarrow a^* + \delta_{a^*}$\;
        ${b^*}' \leftarrow b^* + \delta_{b^*}$\;

        \textbf{$\triangleright$ Convert to RGB space}\;
        $x_{lab} \leftarrow lab2rgb(l^*, {a^*}', {b^*}')$\;

        \textbf{$\triangleright$ Convert to frequency space}\;
        $X \leftarrow DCT(x_{lab})$\;
        $X' \leftarrow X + \delta_{\mathrm{freq}}$\;

        \textbf{$\triangleright$ Convert back to RGB space}\;
        $x_{freq} \leftarrow IDCT(X' \odot M)$\;
        $x_{\xi} \leftarrow clip(x_{freq}, x-\Delta_{RGB}, x+\Delta_{RGB})$\;

        \textbf{$\triangleright$ Update the perturbations}\;
        $\delta_{a^*} \leftarrow \delta_{a^*} - \alpha \cdot \delta_{a^*}\,\mathcal{L}_{\text{Anti-I2V}}(\epsilon_\theta, x_{\xi}, y)$\;
        $\delta_{b^*} \leftarrow \delta_{b^*} - \alpha \cdot \delta_{b^*}\,\mathcal{L}_{\text{Anti-I2V}}(\epsilon_\theta, x_{\xi}, y)$\;
        $\delta_{\mathrm{freq}} \leftarrow \delta_{\mathrm{freq}} - \alpha \cdot \delta_{\mathrm{freq}}\,\mathcal{L}_{\text{Anti-I2V}}(\epsilon_\theta, x_{\xi}, y)$\;
    }
    \Return{$x_{\xi}$}\;
    \caption{Dual-Space Perturbation Optimization}
    \label{alg:procedure}
\end{algorithm}

\begin{figure*}[ht]
    \centering
    \includegraphics[width=\linewidth]{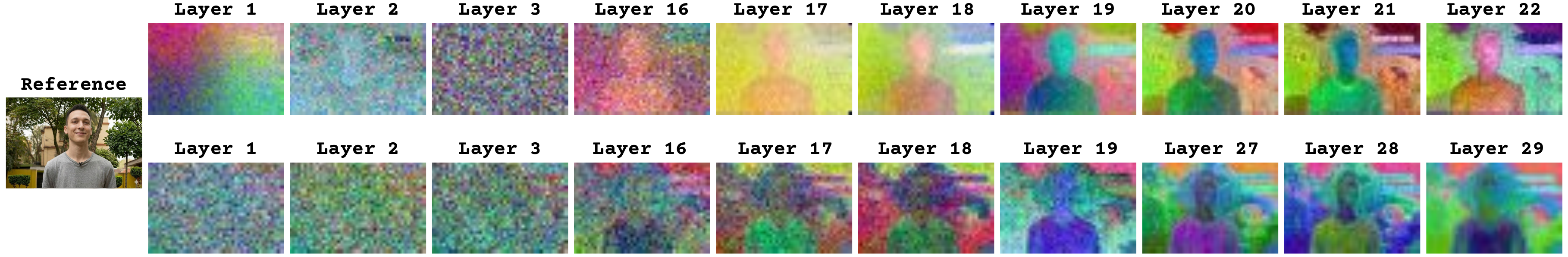} 
    \caption{\textbf{PCA visualization of features in each layer.} Features from each block are visualized at timestep 500. The first row shows features from OpenSora~\cite{opensora}, while the second row shows features from CogVideoX~\cite{cogX}. For clarity, only selected layers are highlighted.}
    \label{fig:feats}
\end{figure*}

\subsection{Internal Representation Collapse Loss} 
\label{subset:IRC}
As mentioned in \cref{subsec:proposal}, to attack large models effectively, perturbations must disrupt feature propagation across layers, preventing the reconstruction of meaningful structures. We achieve this by suppressing rich semantic features into low-information representations. Therefore, it is essential to identify layers that encode rich semantic information and strategically align them with layers that contain minimal semantic content to maximize feature disruption.

\noindent\textbf{Analysis at Different UNet/DiT Layers.} 
Following \cite{plug, simac}, we employ PCA to visualize the output features of each transformer blocks during denoising process of CogVideoX \cite{cogX} and OpenSora \cite{opensora} at timestep 500. Similarly to UNet output features, the visualized features progressively shift from encoding structures and low-frequency details to capturing textures and higher-frequency information. As shown in \cref{fig:feats}, meaningful semantic features emerge after $19^{th}$ layer in OpenSora and $27^{th}$ layer in CogVideoX, whereas the $3^{rd}$ layer contains almost no semantic information in both models. For UNet models, \cite{simac} shows that semantic features emerge in the $6^{th}$ decoder layer, whereas the early $3^{rd}$ layer primarily contains low-semantic features. Based on this observation, we aim to transform the features learned in later layers to resemble those of the early $3^{rd}$ layer. Although the most precise strategy would map all features after the $19^{\text{th}}$ layer of OpenSora and the $27^{\text{th}}$ layer of CogVideoX to their respective $3^{\text{rd}}$ layer, we find that a much simpler layer selection rule achieves comparable performance. We formulate the training loss and selection strategy in the following section.

\noindent\textbf{Internal Representation Collapse (IRC).} Based on the findings in \cref{fig:feats}, we prevent deeper layers from capturing high-level semantic details by aligning their feature maps with those of early layers, which primarily encode structural and low-level features. Specifically, let us denote the set of target deep layers as $J$. For each deep layer $j\in J$ and an early layer $i$, we define the IRC loss as the squared, normalized Euclidean distance between their features:
\begin{equation}
    \mathcal{L}_{\text{IRC}}^{i,j} = \mathbb{E} \left\| \epsilon_{\theta}^j(z_t, z_{\xi}, t, y) - \epsilon_{\theta}^i(z_t, z_{\xi}, t, y) \right\|_2^2,
\end{equation}
where \(z_{t}\) is the noised input at timestep $t$, \(z_{\xi}\) is the latent of the perturbed image conditional input, \(\epsilon_{\theta}^j(.)\) denotes the normalized feature map at the \(j\)-th block during the denoising process, and $y$ is the text embedding. Each feature map is normalized by its map size \(\frac{1}{F C H W}\), where $F$ is the number of video frames. By aligning deep-layer features with those of early layers, we collapse high-level semantic representations in the deeper blocks. As a result, the latent features propagated through the denoising process contain less structural and semantic information, weakening the model’s ability to reconstruct coherent content. Consequently, each denoising step becomes more sensitive to perturbations, amplifying adversarial noise and degrading the fidelity of the generated outputs. In our implementation, setting $j$ to the indices of the last three layers of each model and $i$ to the $3^{\text{rd}}$ layer yields nearly identical protection performance, leading to a simple layer-selection rule in our final objective.

\subsection{Internal Representation Anchor Loss}
\label{subsec:IRA}
While IRC loss focuses on the disruption of information flow within the denoising module, we want to further disrupt the extraction of features contained within each layer of model components. AdvDM \cite{advdm} and MIST \cite{mist} introduce a textual loss that reduces the distance between the encoded representations of the original and perturbed images. However, they focus solely on the final output of VAEs, disregarding the denoising module and their intermediate outputs. Hence, we propose Internal Representation Anchor loss (IRA) that minimizes the layer-wise Euclidean distance between the hidden features produced when conditioning on the perturbed image and on a unrelated target image in both the denoising modules and the VAE \cite{vae}. In this targeted attack setting, IRA is formulated as:
\begin{equation}
    \mathcal{L}_{\text{IRA}, \epsilon_{\theta}}^{m} = \mathbb{E} \left\| \epsilon_{\theta}^m(z_t, z_{\xi}, t, y) - \epsilon_{\theta}^m(z_t, z_{\psi}, t, y) \right\|_2^2
\label{eq:l_ira_epsilon}
\end{equation}
\begin{equation}
    \mathcal{L}_{\text{IRA}, E}^{n} = \mathbb{E} \left\| E^n(z_{\xi}) - E^n(z_{\psi}) \right\|_2^2
\end{equation}
\begin{equation}
    \mathcal{L}_{\text{IRA}} = \mathcal{L}_{\text{IRA}, \epsilon_{\theta}} + \mathcal{L}_{\text{IRA}, E}
\end{equation}
\noindent where $m$ indicates the $m^{th}$ layer of the denoising network, $n$ indicates the $n^{th}$ layer within the VAE, \(z_{\psi}\) is the latent target image, \(z_{\xi}\) is the latent of the perturned conditional image, and $E^n(.)$ represents the encoded latent at the $n^{th}$ VAE layer. In untargeted attacks, \( x_{\psi} = x \), and the sign in \( \mathcal{L}_{\text{IRA}} \) is reversed. A limitation of the textual loss in AdvDM \cite{advdm} and MIST \cite{mist} is that the perturbations adopt textures similar to the target image in the background, making them visually noticeable. However, with DSP, perturbations are introduced in the $L$*$a$*$b$* color and frequency domains, avoiding issues caused by RGB-based modifications.

\subsection{Final Objectives}
\label{subsec:final}
Inspired by \cite{dormant}, we incorporate LPIPS loss and CLIP feature loss. We refer to these collectively as Auxiliary Losses.
\begin{equation}
    \mathcal{L}_{\text{CLIP}} = \left\| \mathcal{C}(x_{\xi}) -\mathcal{C}(x)\right\|_2^2,
\end{equation}
\begin{equation}
    \mathcal{L}_{\text{LPIPS}} = \text{LPIPS}(x_{\xi}, x),
\end{equation}
\begin{equation}
    \mathcal{L}_{\text{auxiliary}} = \mathcal{L}_{\text{CLIP}} - \mathcal{L}_{\text{LPIPS}}
\label{eq:final}
\end{equation}
where $\mathcal{C}$ is the CLIP image encoder and $\mathcal{F}$ is the commonly-used feature extractor in perceptual losses. The final objective function combines four loss components:
\begin{equation}
    \mathcal{L}_{Anti-I2V} = \mathcal{L}_{\text{IRC}} + \mathcal{L}_{\text{IRS}} + \mathcal{L}_{\text{auxiliary}} - \mathcal{L}_{\text{DM}}
\end{equation}

\section{Experiments}\label{sec:exp}
\begin{table*}[!t]
    \centering
    \footnotesize 
    \setlength{\tabcolsep}{3.5pt}
    \caption{\textbf{Quantitative comparison of Anti-I2V against baseline protections.} $\downarrow$ indicates that lower values correspond to poorer video quality and thus stronger protection. The best scores are highlighted in \textbf{bold}, while the second-best results are \underline{underlined}.}
    \begin{tabular}{l l c c c c c c c c c c}
        \toprule
        & & \multicolumn{5}{c}{\textbf{CelebV-Text}} & \multicolumn{5}{c}{\textbf{UCF101}} \\
        \cmidrule(lr){3-7} \cmidrule(lr){8-12}
        \textbf{Model} & \textbf{Method} & \textbf{ISM} $\downarrow$ & \textbf{C-FIQA} $\downarrow$ & \textbf{Q-A(F)} $\downarrow$ & \textbf{Q-A(V)} $\downarrow$ & \textbf{DINO} $\downarrow$ & \textbf{ISM} $\downarrow$ & \textbf{C-FIQA} $\downarrow$ & \textbf{Q-A(F)} $\downarrow$ & \textbf{Q-A(V)} $\downarrow$ & \textbf{DINO} $\downarrow$ \\
        \midrule
        
        \multirow{7}{*}{CogVideoX-5B}
        & Clean & 0.721 & 0.522 & 0.746 & 0.802 & 0.828 & 0.466 & 0.373 & 0.361 & 0.436 & 0.801 \\
        & SDS+ \cite{sds} & 0.591 & 0.482 & 0.473 & 0.563 & 0.754 & 0.381 & 0.291 & 0.286 & 0.344 & 0.754 \\
        & SDS- \cite{sds} & 0.607 & 0.497 & 0.511 & 0.594 & 0.747 & 0.386 & 0.298 & 0.313 & 0.393 & 0.760 \\
        & AdvDM \cite{advdm} & 0.583 & 0.473 & 0.463 & 0.543 & 0.748 & 0.370 & 0.292 & 0.271 & 0.342 & 0.753 \\
        & MIST \cite{mist} & 0.561 & 0.463 & 0.476 & 0.577 & 0.750 & 0.355 & 0.290 & 0.262 & 0.340 & 0.751 \\
        & VGMShield \cite{vgm} & 0.554 & 0.461 & 0.464 & 0.557 & 0.745 & 0.361 & 0.292 & 0.265 & 0.343 & 0.753 \\
        \rowcolor{blue!15} \cellcolor{white} & \textbf{Anti-I2V} & \textbf{0.448} & \textbf{0.433} & \textbf{0.447} & \textbf{0.532} & \textbf{0.722} & \textbf{0.346} & \textbf{0.283} & \textbf{0.251} & \textbf{0.323} & \textbf{0.734} \\
        \midrule
        
        \multirow{7}{*}{DynamiCrafter}
        & Clean & 0.528 & 0.467 & 0.724 & 0.794 & 0.622 & 0.384 & 0.345 & 0.501 & 0.562 & 0.709 \\
        & SDS+ \cite{sds} & 0.264 & 0.372 & 0.213 & 0.245 & 0.389 & 0.122 & 0.305 & 0.157 & 0.194 & 0.433 \\
        & SDS- \cite{sds} & 0.278 & 0.418 & 0.223 & 0.250 & 0.392 & 0.128 & 0.338 & 0.164 & 0.226 & 0.439 \\
        & AdvDM \cite{advdm} & 0.269 & 0.370 & 0.167 & 0.207 & 0.397 & 0.110 & 0.335 & 0.162 & 0.201 & 0.451 \\
        & MIST \cite{mist} & 0.262 & 0.379 & 0.232 & 0.269 & 0.386 & 0.100 & 0.335 & 0.322 & 0.381 & 0.493 \\
        & VGMShield \cite{vgm} & 0.286 & 0.431 & 0.243 & 0.289 & 0.401 & 0.108 & 0.336 & 0.318 & 0.374 & 0.486 \\
        \rowcolor{blue!15} \cellcolor{white} & \textbf{Anti-I2V} & \textbf{0.151} & \textbf{0.303} & \textbf{0.032} & \textbf{0.047} & \textbf{0.167} & \textbf{0.068} & \textbf{0.268} & \textbf{0.057} & \textbf{0.084} & \textbf{0.164} \\
        \midrule

        \multirow{7}{*}{Open-Sora}
        & Clean & 0.598 & 0.508 & 0.712 & 0.782 & 0.811 & 0.400 & 0.382 & 0.409 & 0.437 & 0.750 \\
        & SDS+ \cite{sds} & 0.502 & 0.481 & 0.494 & \textbf{0.548} & 0.730 & 0.355 & 0.293 & 0.360 & 0.389 & 0.692 \\
        & SDS- \cite{sds}& 0.508 & 0.494 & 0.514 & 0.591 & 0.731 & 0.333 & 0.311 & 0.351 & 0.387 & 0.687 \\
        & AdvDM \cite{advdm} & 0.506 & 0.478 & 0.496 & 0.561 & 0.725 & 0.346 & 0.309 & 0.327 & 0.362 & 0.686 \\
        & MIST \cite{mist} & 0.493 & 0.475 & 0.497 & 0.594 & \textbf{0.710} & 0.339 & 0.309 & 0.338 & 0.392 & 0.677 \\
        & VGMShield \cite{vgm} & 0.500 & 0.476 & 0.497 & 0.578 & 0.716 & 0.341 & 0.312 & 0.335 & 0.369 & 0.680 \\
        \rowcolor{blue!15} \cellcolor{white} & \textbf{Anti-I2V} & \textbf{0.461} & \textbf{0.453} & \textbf{0.478} & \underline{0.554} & \underline{0.713} & \textbf{0.318} & \textbf{0.248} & \textbf{0.311} & \textbf{0.347} & \textbf{0.642} \\
        \bottomrule
    \end{tabular}
    \label{tab:main_exp}
\end{table*}

\begin{figure*}[ht]
    \centering
    \includegraphics[width=\linewidth]{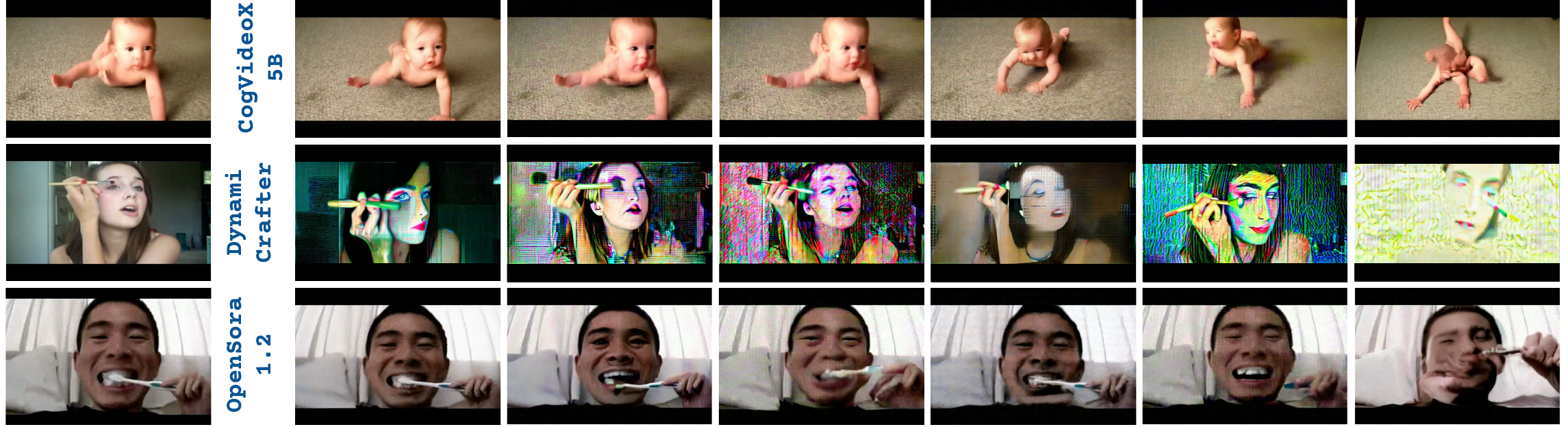} 
    \caption{\textbf{Quanlitative comparison of Anti-I2V against baseline protections against different video generation models on UCF101.} The columns present the generated outputs from the models under different adversarial attack methods.}
    \vspace{-5mm}
    \label{fig:quali}
\end{figure*}

\subsection{Evaluation Dataset}
Although several I2V benchmarks exist, none are specifically designed for both face-centric and human-centric animation~\cite{vbench,tip,tiktok}. Therefore, we establish two evaluation protocols based on complementary datasets.

\noindent\textbf{CelebV-Text.} For face-centric video synthesis, we build a benchmark based on CelebV-Text~\cite{celebvtext}. We select 1,000 videos with unique identities and generate 5,000 videos in total. For convenience, we refer to this benchmark as \textit{CelebV-Text}, designed for face-centric video synthesis.

\noindent\textbf{UCF101.} Since our CelebV-Text mainly contains close-up interview-style videos, we additionally evaluate dynamic full-body actions based on UCF101~\cite{ucf}. We obtain action descriptions, randomly select 200 videos across diverse classes, and generate five samples per pair, resulting in 1,000 total videos. We refer to this benchmark as \textit{UCF101}, designed for full-body human actions synthesis.

\subsection{Experimental Setup}
\noindent \textbf{Models.} We comprehensively evaluate each method on three widely used video diffusion models: CogVideoX-5B I2V \cite{cogX}, OpenSora v1.2 \cite{opensora}, and DynamiCrafter \cite{dynamic}. These models represent state-of-the-art open-source image-conditioned text-to-video generation and cover major architectural types: MMDiT-based, DiT-based and U-Net-based.

\noindent \textbf{Baselines.} We compare several open-source adversarial attack methods for protecting user images against diffusion models, including SDS \cite{sds}, AdvDM \cite{advdm}, MIST \cite{mist}, and VGMShield \cite{vgm}. Since the target video diffusion model is not finetuned, we skip methods with surrogate models \cite{antidb, simac}. We exclude DORMANT~\cite{dormant}, Vid-Freeze~\cite{vidfreeze}, and I2VGuard~\cite{i2vguard} due to their reliance on reference poses, high-end GPUs, or the absence of public implementations. For a fair comparison, we set the perturbation budget in both RGB and $L^*a^*b^*$ space as $16/255$ and iterations $N=200$ for all methods. In Anti-I2V, we set $M$ to keep the top 25\% low frequency components during optimization. All experiments are conducted on a single NVIDIA A100 40GB GPU.

\noindent\textbf{Metric.} We measure Identity Score Matching (ISM) by extracting ArcFace \cite{arcface} embeddings from detected faces and computing the cosine distance to the clean reference. We also use CLIP-FIQA (C-FIQA) \cite{fiqa}, a recent advanced metric specifically designed for facial images. For video quality assessment, we use Q-Align \cite{qlign}, a recent state-of-the-art image and video evaluation metric. We compute Q-Align scores for individual frames and entire videos, denoted as Q-A(F) and Q-A(V), respectively. Additionally, DINO denotes cosine similarity between DINO-extracted features \cite{dino} of generated frames and clean reference image.
\begin{table*}[!t]
    \centering
    \footnotesize
    \setlength{\tabcolsep}{3.5pt} 
    \caption{\textbf{Quantitative comparison of Anti-I2V against baseline video protection methods.} Lower values indicate stronger protection (poorer video quality). Best scores are \textbf{bold}, second-best are \underline{underlined}.}
    \begin{tabular}{l c c c c c c c c c c}
        \toprule
        & \multicolumn{5}{c}{\textbf{CogVideoX-5B \cite{cogX} - OpenSora 1.2 \cite{opensora}}} & \multicolumn{5}{c}{\textbf{CogVideoX-5B \cite{cogX} - DynamiCrafter \cite{dynamic}}} \\
        \cmidrule(lr){2-6} \cmidrule(lr){7-11}
        \textbf{Method} & ISM $\downarrow$ & C-FIQA $\downarrow$ & Q-A(F) $\downarrow$ & Q-A(V) $\downarrow$ & DINO $\downarrow$ 
                         & ISM $\downarrow$ & C-FIQA $\downarrow$ & Q-A(F) $\downarrow$ & Q-A(V) $\downarrow$ & DINO $\downarrow$ \\
        \midrule
        SDS+ \cite{sds}       & 0.606 & 0.493 & 0.514 & \textbf{0.582} & 0.764 & 0.309 & 0.428 & 0.432 & 0.519 & \textbf{0.614} \\
        SDS- \cite{sds}       & 0.568 & 0.504 & 0.563 & 0.641 & 0.757 & \textbf{0.286} & 0.422 & 0.493 & 0.579 & 0.664 \\
        AdvDM \cite{advdm}    & 0.608 & 0.491 & 0.506 & 0.593 & 0.754 & 0.337 & 0.428 & 0.389 & 0.511 & 0.631 \\
        MIST \cite{mist}      & 0.547 & 0.490 & 0.522 & 0.636 & \textbf{0.726} & \underline{0.289} & 0.440 & 0.384 & 0.518 & 0.633 \\
        VGMShield \cite{vgm}  & 0.566 & 0.503 & 0.548 & 0.632 & 0.751 & 0.322 & 0.462 & 0.387 & 0.504 & 0.688 \\
        \rowcolor{blue!15} \textbf{Anti-I2V} 
                               & \textbf{0.467} & \textbf{0.476} & \textbf{0.501} & \underline{0.588} & \underline{0.736} 
                               & 0.294 & 0.431 & \textbf{0.376} & \textbf{0.491} & \underline{0.620} \\
        \bottomrule
    \end{tabular}
    \label{tab:transfer}
\end{table*}
\subsection{Quantitative Results}
\cref{tab:main_exp} reports the performance of Anti-I2V against various baseline protection methods across three video generation models. On both CelebV-Text and UCF101, Anti-I2V consistently achieves the lowest ISM and CLIP-FIQA, indicating a substantial reduction in identity similarity between generated and reference faces. For instance, on the UNet-based DynamiCrafter, Anti-I2V achieves an ISM of \textbf{0.151} and Q-Align (V) of \textbf{0.047}, significant drops from the clean outputs (0.528 and 0.794, respectively) and large improvements over the next best baselines (0.262 and 0.207). Similarly, it achieves the lowest CLIP-FIQA of \textbf{0.303}, further indicating degraded facial fidelity. On UCF101, Anti-I2V again yields the lowest ISM of \textbf{0.068} and Q-Align (V) of \textbf{0.084}, compared to 0.384 and 0.562 for clean videos. Reductions in DINO scores, Q-Align (V), and Q-Align (F) across models indicate a deterioration in temporal coherence and overall visual quality. The consistently strong results across both datasets and models demonstrate that Anti-I2V effectively disrupts facial identity preservation, frame-to-frame consistency, and motion smoothness, resulting in significantly less natural video synthesis.

\subsection{Qualitative Results}
\noindent\cref{fig:quali} shows that Anti-I2V outperforms other baselines in both identity degradation and overall video quality reduction across multiple models. While competing methods mainly introduce minor blurring and color shifts, Anti-I2V effectively disrupts identity features and produces pronounced artifacts. In CogVideoX and Open-Sora~1.2, baseline methods fail to cause significant identity degradation, whereas Anti-I2V alters facial features and introduces strong distortions throughout the video. In DynamiCrafter, Anti-I2V further induces severe color distortions that disrupt visual coherence, rendering outputs unrecognizable.

\section{Ablation}
Ablation experiments are conducted using CogVideoX-5B \cite{cogX} on a \textbf{subset of 200 videos} from our CelebV-Text \cite{celebvtext}. For each image–prompt pair, we generate five samples, resulting in \textbf{1,000 videos} for evaluation.

\subsection{Robustness}
\label{sec:robust}
We evaluate the robustness of our Dual-Space Perturbation and RGB perturbation under three transformations: JPEG compression, Gaussian blur, and Gaussian noise. Additionally, we benchmark both approaches against purification techniques, DiffPure \cite{diffpure}, GrIDPure \cite{gridpure}, and Impress \cite{impress}. All experiments use the same objective function, $\mathcal{L}_{Anti-I2V}$. \noindent \cref{tab:robustness} shows that DSP achieves more stable performance than RGB, reducing variability and improving robustness under JPEG compression, Gaussian blur, DiffPure, Impress, and Gaussian noise.

\begin{table}[!t]
\caption{\textbf{Quantitative results against various transformations and purifications.} \green{Green} indicates degrees of improvements, while \red{Red} indicates degrees of performance drop.}
\label{tab:robustness}
\centering
\scriptsize
\tabcolsep=0.05cm
\renewcommand\arraystretch{1}
\begin{tabular}{lcccc}
    \toprule
    \midrule
    {\textbf{Method}} &\textbf{ISM$\downarrow$} &\textbf{Q-A (F)$\downarrow$} &\textbf{Q-A (V)$\downarrow$} &\textbf{DINO-SIM$\downarrow$} \\
    \midrule
    \textbf{Ours (RGB)} & \textbf{0.55} & \textbf{0.50} & \textbf{0.58} & \textbf{0.76}\\
    \midrule
    + DiffPure \cite{diffpure} & 0.33 \green{(-0.22)} & 0.49 \green{(-0.01)} & 0.59 \red{(+0.01)} & 0.74 \green{(-0.02)} \\
    + GrIDPure \cite{gridpure} & 0.60 \red{(+0.05)} & 0.70 \red{(+0.20)} & 0.75 \red{(+0.17)} & 0.84 \red{(+0.08)} \\
    + Impress \cite{impress} & 0.58 \red{(+0.03)} & 0.52 \red{(+0.02)} & 0.61 \red{(+0.03)} & \textbf{0.76 \green{(-0.00)}} \\
    + JPEG Compression & 0.57 \red{(+0.02)} & 0.51 \red{(+0.01)} & 0.58 \red{(-0.00)} & 0.78 \red{(+0.02)} \\
    + Gaussian Blur & 0.63 \red{(+0.08)} & 0.43 \green{(-0.07)} & 0.51 \green{(-0.07)} & 0.80 \red{(+0.04)} \\
    + Gaussian Noise & 0.58 \red{(+0.03)} & 0.50 \red{(-0.00)} & 0.58 \red{(-0.00)} & 0.78 \red{(+0.02)} \\
    \bottomrule	
\end{tabular}
\centering
\scriptsize
\tabcolsep=0.05cm
\renewcommand\arraystretch{1}
\begin{tabular}{lcccc}
    \toprule
    {\textbf{Method}} &\textbf{ISM$\downarrow$} &\textbf{Q-A (F)$\downarrow$} &\textbf{Q-A (V)$\downarrow$} &\textbf{DINO-SIM$\downarrow$} \\
    \midrule
    \textbf{Ours (DSP)} & \textbf{0.46} & \textbf{0.48} & \textbf{0.56} & \textbf{0.76}\\
    \midrule
    + DiffPure \cite{diffpure} & 0.33 \green{(-0.13)} & 0.44 \green{(-0.04)} & 0.55 \green{(-0.01)} & 0.73 \green{(-0.03)} \\
    + GrIDPure \cite{gridpure} & 0.49 \red{(+0.03)} & 0.58 \red{(+0.10)} & 0.68 \red{(+0.12)} & 0.80 \red{(+0.04)} \\
    + Impress \cite{impress} & \textbf{0.46 \green{(-0.00)}} & 0.49 \red{(+0.01)} & 0.57 \red{(+0.01)} & \textbf{0.76 \green{(-0.00)}} \\
    + JPEG Compression & 0.47 \red{(+0.01)} & 0.44 \green{(-0.04)} & 0.54 \green{(-0.02)} & 0.77 \red{(+0.01)}\\
    + Gaussian Blur & 0.49 \red{(+0.03)} & 0.32 \green{(-0.16)} & 0.41 \green{(-0.15)} & 0.77 \red{(+0.01)} \\
    + Gaussian Noise & 0.49 \red{(+0.03)} & 0.42 \green{(-0.06)} & 0.52 \green{(-0.04)} & 0.77 \red{(+0.01)} \\
    \midrule
    \bottomrule	
\end{tabular}
\end{table}

\subsection{Transferability}
\label{sec:transfer}
We evaluate the transferability of our method across diffusion models under two settings: MMDiT–DiT and MMDiT–UNet transfer. For DiT-based transfer, adversarial images are optimized on CogVideoX \cite{cogX} and evaluated on OpenSora~1.2~\cite{opensora}. For cross-architecture transfer, we optimize on CogVideoX and test on DynamiCrafter~\cite{dynamic}. Since each model requires different input sizes, adversarial images are resized accordingly. As shown in \cref{tab:transfer}, our method achieves the strongest identity degradation in the DiT-based setting, obtaining the best ISM and CLIP-FIQA scores while maintaining comparable video quality to baselines. In the DiT–UNet setting, it achieves the best overall performance albeit exhibiting performance degradation.

\section{Conclusion and Future Works}
We propose Anti-I2V, a novel approach to prevent unauthorized image usage in text-image-to-video generation. By applying protective perturbations in the $L^*a^*b^*$ color space and frequency domain, Anti-I2V embeds robust disruptions beyond raw RGB pixel intensities. Moreover, we introduce novel $\mathcal{L}_{Anti-I2V}$ to disrupt information flow and degrade hidden features across network layers. Experimental results demonstrate the effectiveness of Anti-I2V, establishing it as a strong defense against this security threat.
\clearpage
{
    \small
    \bibliographystyle{ieeenat_fullname}
    \bibliography{main}

@String(CVPR= {IEEE Conf. Comput. Vis. Pattern Recog.})

@String(ECCV= {Eur. Conf. Comput. Vis.})

@String(TIP  = {IEEE Trans. Image Process.})

@String(ICLR = {Int. Conf. Learn. Represent.})

@String(CVPR  = {CVPR})

@String(ECCV  = {ECCV})

@String(TIP   = {IEEE TIP})

@String(ICLR  = {ICLR})

@article{sora,
  title={Video generation models as world simulators. 2024},
  author={Brooks, Tim and Peebles, Bill and Holmes, Connor and DePue, Will and Guo, Yufei and Jing, Li and Schnurr, David and Taylor, Joe and Luhman, Troy and Luhman, Eric and others},
  journal={URL https://openai. com/research/video-generation-models-as-world-simulators},
  volume={3},
  year=2024
}

@article{gen3,
  title={Introducing Gen-3 Alpha: A New Frontier for Video Generation. 2024},
  author={Germanidis, Anastasis},
  journal={https://runwayml.com/research/introducing-gen-3-alpha},
  year=2024
}

@article{kling,
  title={KLING, MAKE IMAGINATION ALIVE. 2024},
  author={Kling AI},
  journal={https://klingai.io/},
  year=2024
}

@article{svd,
  title={Stable video diffusion: Scaling latent video diffusion models to large datasets},
  author={Blattmann, Andreas and Dockhorn, Tim and Kulal, Sumith and Mendelevitch, Daniel and Kilian, Maciej and Lorenz, Dominik and Levi, Yam and English, Zion and Voleti, Vikram and Letts, Adam and others},
  journal={arXiv preprint arXiv:2311.15127},
  year={2023}
}

@article{vcrafter1,
  title={VideoCrafter1: Open Diffusion Models for High-Quality Video Generation},
  author={Haoxin Chen and Menghan Xia and Yin-Yin He and Yong Zhang and Xiaodong Cun and Shaoshu Yang and Jinbo Xing and Yaofang Liu and Qifeng Chen and Xintao Wang and Chao-Liang Weng and Ying Shan},
  journal={ArXiv},
  year={2023},
  volume={abs/2310.19512}
}

@article{vcrafter2,
  title={VideoCrafter2: Overcoming Data Limitations for High-Quality Video Diffusion Models},
  author={Haoxin Chen and Yong Zhang and Xiaodong Cun and Menghan Xia and Xintao Wang and Chao-Liang Weng and Ying Shan},
  journal={2024 IEEE/CVF Conference on Computer Vision and Pattern Recognition (CVPR)},
  year={2024},
  pages={7310-7320}
}

@article{modelscope,
  title={ModelScope Text-to-Video Technical Report},
  author={Jiuniu Wang and Hangjie Yuan and Dayou Chen and Yingya Zhang and Xiang Wang and Shiwei Zhang},
  journal={ArXiv},
  year={2023},
  volume={abs/2308.06571}
}

@article{cog,
  title={Cogvideo: Large-scale pretraining for text-to-video generation via transformers},
  author={Hong, Wenyi and Ding, Ming and Zheng, Wendi and Liu, Xinghan and Tang, Jie},
  journal={arXiv preprint arXiv:2205.15868},
  year={2022}
}

@article{cogX,
  title={CogVideo: Large-scale Pretraining for Text-to-Video Generation via Transformers},
  author={Hong, Wenyi and Ding, Ming and Zheng, Wendi and Liu, Xinghan and Tang, Jie},
  journal={arXiv preprint arXiv:2205.15868},
  year={2022}
}

@software{opensora,
  author = {Zangwei Zheng and Xiangyu Peng and Tianji Yang and Chenhui Shen and Shenggui Li and Hongxin Liu and Yukun Zhou and Tianyi Li and Yang You},
  title = {Open-Sora: Democratizing Efficient Video Production for All},
  month = {March},
  year = {2024},
  url = {https://github.com/hpcaitech/Open-Sora}
}

@article{i2vgen,
  title={I2vgen-xl: High-quality image-to-video synthesis via cascaded diffusion models},
  author={Zhang, Shiwei and Wang, Jiayu and Zhang, Yingya and Zhao, Kang and Yuan, Hangjie and Qin, Zhiwu and Wang, Xiang and Zhao, Deli and Zhou, Jingren},
  journal={arXiv preprint arXiv:2311.04145},
  year={2023}
}

@inproceedings{animatediff,
  title={AnimateDiff: Animate Your Personalized Text-to-Image Diffusion Models without Specific Tuning},
  author={Guo, Yuwei and Yang, Ceyuan and Rao, Anyi and Liang, Zhengyang and Wang, Yaohui and Qiao, Yu and Agrawala, Maneesh and Lin, Dahua and Dai, Bo},
  booktitle={The Twelfth International Conference on Learning Representations}
}

@article{animatelcm,
  title={Animatelcm: Accelerating the animation of personalized diffusion models and adapters with decoupled consistency learning},
  author={Wang, Fu-Yun and Huang, Zhaoyang and Shi, Xiaoyu and Bian, Weikang and Song, Guanglu and Liu, Yu and Li, Hongsheng},
  journal={arXiv preprint arXiv:2402.00769},
  year={2024}
}

@inproceedings{advdm,
  title={Adversarial example does good: preventing painting imitation from diffusion models via adversarial examples},
  author={Liang, Chumeng and Wu, Xiaoyu and Hua, Yang and Zhang, Jiaru and Xue, Yiming and Song, Tao and Xue, Zhengui and Ma, Ruhui and Guan, Haibing},
  booktitle={Proceedings of the 40th International Conference on Machine Learning},
  pages={20763--20786},
  year={2023}
}

@inproceedings{antidb,
  title={Anti-dreambooth: Protecting users from personalized text-to-image synthesis},
  author={Van Le, Thanh and Phung, Hao and Nguyen, Thuan Hoang and Dao, Quan and Tran, Ngoc N and Tran, Anh},
  booktitle={Proceedings of the IEEE/CVF International Conference on Computer Vision},
  pages={2116--2127},
  year={2023}
}

@article{mist,
  title={Mist: Towards Improved Adversarial Examples for Diffusion Models},
  author={Liang, Chumeng and Wu, Xiaoyu},
  journal={arXiv preprint arXiv:2305.12683},
  year={2023}
}

@inproceedings{glaze,
  title={Glaze: protecting artists from style mimicry by text-to-image models},
  author={Shan, Shawn and Cryan, Jenn and Wenger, Emily and Zheng, Haitao and Hanocka, Rana and Zhao, Ben Y},
  booktitle={Proceedings of the USENIX conference},
  year={2023},
  organization={USENIX Association}
}

@inproceedings{duaw,
  title={DUAW: Data-free Universal Adversarial Watermark against Stable Diffusion Customization},
  author={Ye, Xiaoyu and Huang, Hao and An, Jiaqi and Wang, Yongtao},
  booktitle={ICLR 2024 Workshop on Secure and Trustworthy Large Language Models},

}

@inproceedings{sds,
  title={Toward effective protection against diffusion-based mimicry through score distillation},
  author={Xue, Haotian and Liang, Chumeng and Wu, Xiaoyu and Chen, Yongxin},
  booktitle={The Twelfth International Conference on Learning Representations}
}

@misc{vgm,
      title={VGMShield: Mitigating Misuse of Video Generative Models}, 
      author={Yan Pang and Yang Zhang and Tianhao Wang},
      year={2024},
      eprint={2402.13126},
      archivePrefix={arXiv},
      primaryClass={cs.CR}
}

@article{dormant,
  title={Dormant: Defending against Pose-driven Human Image Animation},
  author={Jiachen Zhou and Mingsi Wang and Tianlin Li and Guozhu Meng and Kai Chen},
  journal={ArXiv},
  year={2024},
  volume={abs/2409.14424},
  url={https://api.semanticscholar.org/CorpusID:272827839}
}

@inproceedings{clip,
  title={Learning transferable visual models from natural language supervision},
  author={Radford, Alec and Kim, Jong Wook and Hallacy, Chris and Ramesh, Aditya and Goh, Gabriel and Agarwal, Sandhini and Sastry, Girish and Askell, Amanda and Mishkin, Pamela and Clark, Jack and others},
  booktitle={International conference on machine learning},
  pages={8748--8763},
  year={2021},
  organization={PmLR}
}

@inproceedings{animateanyone,
  title={Animate anyone: Consistent and controllable image-to-video synthesis for character animation},
  author={Hu, Li},
  booktitle={Proceedings of the IEEE/CVF Conference on Computer Vision and Pattern Recognition},
  pages={8153--8163},
  year={2024}
}

@inproceedings{magicanimate,
  title={Magicanimate: Temporally consistent human image animation using diffusion model},
  author={Xu, Zhongcong and Zhang, Jianfeng and Liew, Jun Hao and Yan, Hanshu and Liu, Jia-Wei and Zhang, Chenxu and Feng, Jiashi and Shou, Mike Zheng},
  booktitle={Proceedings of the IEEE/CVF Conference on Computer Vision and Pattern Recognition},
  pages={1481--1490},
  year={2024}
}

@misc{flux,
    author={Black Forest Labs},
    title={FLUX},
    year={2024},
    howpublished={\url{https://github.com/black-forest-labs/flux}},
}

@article{ddpm,
  title={Denoising diffusion probabilistic models},
  author={Ho, Jonathan and Jain, Ajay and Abbeel, Pieter},
  journal={Advances in neural information processing systems},
  volume={33},
  pages={6840--6851},
  year={2020}
}

@inproceedings{ldm,
  title={High-resolution image synthesis with latent diffusion models},
  author={Rombach, Robin and Blattmann, Andreas and Lorenz, Dominik and Esser, Patrick and Ommer, Bj{\"o}rn},
  booktitle={Proceedings of the IEEE/CVF conference on computer vision and pattern recognition},
  pages={10684--10695},
  year={2022}
}

@inproceedings{sdxl,
  title={SDXL: Improving Latent Diffusion Models for High-Resolution Image Synthesis},
  author={Podell, Dustin and English, Zion and Lacey, Kyle and Blattmann, Andreas and Dockhorn, Tim and M{\"u}ller, Jonas and Penna, Joe and Rombach, Robin},
  booktitle={The Twelfth International Conference on Learning Representations}
}

@article{vdm,
  title={Video diffusion models},
  author={Ho, Jonathan and Salimans, Tim and Gritsenko, Alexey and Chan, William and Norouzi, Mohammad and Fleet, David J},
  journal={Advances in Neural Information Processing Systems},
  volume={35},
  pages={8633--8646},
  year={2022}
}

@inproceedings{align,
  title={Align your latents: High-resolution video synthesis with latent diffusion models},
  author={Blattmann, Andreas and Rombach, Robin and Ling, Huan and Dockhorn, Tim and Kim, Seung Wook and Fidler, Sanja and Kreis, Karsten},
  booktitle={Proceedings of the IEEE/CVF conference on computer vision and pattern recognition},
  pages={22563--22575},
  year={2023}
}

@article{show1,
  title={Show-1: Marrying pixel and latent diffusion models for text-to-video generation},
  author={Zhang, David Junhao and Wu, Jay Zhangjie and Liu, Jia-Wei and Zhao, Rui and Ran, Lingmin and Gu, Yuchao and Gao, Difei and Shou, Mike Zheng},
  journal={International Journal of Computer Vision},
  pages={1--15},
  year={2024},
  publisher={Springer}
}

@article{soraplan,
  title={Open-sora plan: Open-source large video generation model},
  author={Lin, Bin and Ge, Yunyang and Cheng, Xinhua and Li, Zongjian and Zhu, Bin and Wang, Shaodong and He, Xianyi and Ye, Yang and Yuan, Shenghai and Chen, Liuhan and others},
  journal={arXiv preprint arXiv:2412.00131},
  year={2024}
}

@inproceedings{vae,
  title={Taming transformers for high-resolution image synthesis},
  author={Esser, Patrick and Rombach, Robin and Ommer, Bjorn},
  booktitle={Proceedings of the IEEE/CVF conference on computer vision and pattern recognition},
  pages={12873--12883},
  year={2021}
}

@inproceedings{dynamic,
  title={Dynamicrafter: Animating open-domain images with video diffusion priors},
  author={Xing, Jinbo and Xia, Menghan and Zhang, Yong and Chen, Haoxin and Yu, Wangbo and Liu, Hanyuan and Liu, Gongye and Wang, Xintao and Shan, Ying and Wong, Tien-Tsin},
  booktitle={European Conference on Computer Vision},
  pages={399--417},
  year={2024},
  organization={Springer}
}

@article{ltx,
  title={Ltx-video: Realtime video latent diffusion},
  author={HaCohen, Yoav and Chiprut, Nisan and Brazowski, Benny and Shalem, Daniel and Moshe, Dudu and Richardson, Eitan and Levin, Eran and Shiran, Guy and Zabari, Nir and Gordon, Ori and others},
  journal={arXiv preprint arXiv:2501.00103},
  year={2024}
}

@article{antiforgery,
  title={Anti-forgery: Towards a stealthy and robust deepfake disruption attack via adversarial perceptual-aware perturbations},
  author={Wang, Run and Huang, Ziheng and Chen, Zhikai and Liu, Li and Chen, Jing and Wang, Lina},
  journal={arXiv preprint arXiv:2206.00477},
  year={2022}
}

@inproceedings{inmark,
  title={Countering personalized text-to-image generation with influence watermarks},
  author={Liu, Hanwen and Sun, Zhicheng and Mu, Yadong},
  booktitle={Proceedings of the IEEE/CVF Conference on Computer Vision and Pattern Recognition},
  pages={12257--12267},
  year={2024}
}

@article{dct,
  title={Discrete cosine transform},
  author={Ahmed, Nasir and Natarajan, T\_ and Rao, Kamisetty R},
  journal={IEEE transactions on Computers},
  volume={100},
  number={1},
  pages={90--93},
  year={2006},
  publisher={IEEE}
}

@inproceedings{hfdb,
  title={High-Frequency Anti-DreamBooth: Robust Defense against Personalized Image Synthesis},
  author={Onikubo, Takuto and Matsui, Yusuke},
  booktitle={ECCV 2024 Workshop The Dark Side of Generative AIs and Beyond}
}

@InProceedings{simac,
    title     = {SimAC: A Simple Anti-Customization Method for Protecting Face Privacy against Text-to-Image Synthesis of Diffusion Models},
    booktitle = {Proceedings of the IEEE/CVF Conference on Computer Vision and Pattern Recognition (CVPR)},
    month     = {June},
    year      = {2024},
    pages     = {12047-12056}
}

@InProceedings{plug,
    author    = {Tumanyan, Narek and Geyer, Michal and Bagon, Shai and Dekel, Tali},
    title     = {Plug-and-Play Diffusion Features for Text-Driven Image-to-Image Translation},
    booktitle = {Proceedings of the IEEE/CVF Conference on Computer Vision and Pattern Recognition (CVPR)},
    month     = {June},
    year      = {2023},
    pages     = {1921-1930}
}

@article{sds3d,
  title={Dreamfusion: Text-to-3d using 2d diffusion},
  author={Poole, Ben and Jain, Ajay and Barron, Jonathan T and Mildenhall, Ben},
  journal={arXiv preprint arXiv:2209.14988},
  year={2022}
}

@inproceedings{celebvtext,
  title={Celebv-text: A large-scale facial text-video dataset},
  author={Yu, Jianhui and Zhu, Hao and Jiang, Liming and Loy, Chen Change and Cai, Weidong and Wu, Wayne},
  booktitle={Proceedings of the IEEE/CVF Conference on Computer Vision and Pattern Recognition},
  pages={14805--14814},
  year={2023}
}

@inproceedings{arcface,
  title={Arcface: Additive angular margin loss for deep face recognition},
  author={Deng, Jiankang and Guo, Jia and Xue, Niannan and Zafeiriou, Stefanos},
  booktitle={Proceedings of the IEEE/CVF conference on computer vision and pattern recognition},
  pages={4690--4699},
  year={2019}
}

@inproceedings{fawkes,
  title={Fawkes: Protecting privacy against unauthorized deep learning models},
  author={Shan, Shawn and Wenger, Emily and Zhang, Jiayun and Li, Huiying and Zheng, Haitao and Zhao, Ben Y},
  booktitle={29th USENIX security symposium (USENIX Security 20)},
  pages={1589--1604},
  year={2020}
}

@inproceedings{qlign,
  title={Q-ALIGN: teaching LMMs for visual scoring via discrete text-defined levels},
  author={Wu, Haoning and Zhang, Zicheng and Zhang, Weixia and Chen, Chaofeng and Liao, Liang and Li, Chunyi and Gao, Yixuan and Wang, Annan and Zhang, Erli and Sun, Wenxiu and others},
  booktitle={Proceedings of the 41st International Conference on Machine Learning},
  pages={54015--54029},
  year={2024}
}

@inproceedings{pgd,
  author       = {Aleksander Madry and
                  Aleksandar Makelov and
                  Ludwig Schmidt and
                  Dimitris Tsipras and
                  Adrian Vladu},
  title        = {Towards Deep Learning Models Resistant to Adversarial Attacks},
  booktitle    = {6th International Conference on Learning Representations, {ICLR} 2018},
}

@inproceedings{vbench,
  title={Vbench: Comprehensive benchmark suite for video generative models},
  author={Huang, Ziqi and He, Yinan and Yu, Jiashuo and Zhang, Fan and Si, Chenyang and Jiang, Yuming and Zhang, Yuanhan and Wu, Tianxing and Jin, Qingyang and Chanpaisit, Nattapol and others},
  booktitle={Proceedings of the IEEE/CVF Conference on Computer Vision and Pattern Recognition},
  pages={21807--21818},
  year={2024}
}

@article{tip,
  title={TIP-I2V: A Million-Scale Real Text and Image Prompt Dataset for Image-to-Video Generation},
  author={Wang, Wenhao and Yang, Yi},
  journal={arXiv preprint arXiv:2411.04709},
  year={2024}
}

@inproceedings{tiktok,
  title={Learning high fidelity depths of dressed humans by watching social media dance videos},
  author={Jafarian, Yasamin and Park, Hyun Soo},
  booktitle={Proceedings of the IEEE/CVF Conference on Computer Vision and Pattern Recognition},
  pages={12753--12762},
  year={2021}
}

@inproceedings{diffpure,
  title={Diffusion Models for Adversarial Purification},
  author={Nie, Weili and Guo, Brandon and Huang, Yujia and Xiao, Chaowei and Vahdat, Arash and Anandkumar, Anima},
  booktitle = {International Conference on Machine Learning (ICML)},
  year={2022}
}

@inproceedings{gridpure,
  title={Can protective perturbation safeguard personal data from being exploited by stable diffusion?},
  author={Zhao, Zhengyue and Duan, Jinhao and Xu, Kaidi and Wang, Chenan and Zhang, Rui and Du, Zidong and Guo, Qi and Hu, Xing},
  booktitle={Proceedings of the IEEE/CVF Conference on Computer Vision and Pattern Recognition},
  pages={24398--24407},
  year={2024}
}

@article{ucf,
  title={UCF101: A dataset of 101 human actions classes from videos in the wild},
  author={Soomro, Khurram and Zamir, Amir Roshan and Shah, Mubarak},
  journal={arXiv preprint arXiv:1212.0402},
  year={2012}
}

@article{impress,
  title={Impress: Evaluating the resilience of imperceptible perturbations against unauthorized data usage in diffusion-based generative ai},
  author={Cao, Bochuan and Li, Changjiang and Wang, Ting and Jia, Jinyuan and Li, Bo and Chen, Jinghui},
  journal={Advances in Neural Information Processing Systems},
  volume={36},
  pages={10657--10677},
  year={2023}
}

@inproceedings{i2vguard,
  title={I2VGuard: Safeguarding Images against Misuse in Diffusion-based Image-to-Video Models},
  author={Gui, Dongnan and Guo, Xun and Zhou, Wengang and Lu, Yan},
  booktitle={Proceedings of the Computer Vision and Pattern Recognition Conference},
  pages={12595--12604},
  year={2025}
}

@article{vidfreeze,
  title={Vid-Freeze: Protecting Images from Malicious Image-to-Video Generation via Temporal Freezing},
  author={Chowdhury, Rohit and Bala, Aniruddha and Jaiswal, Rohan and Roheda, Siddharth},
  journal={arXiv preprint arXiv:2509.23279},
  year={2025}
}

@inproceedings{fiqa,
  title={Clib-fiqa: Face image quality assessment with confidence calibration},
  author={Ou, Fu-Zhao and Li, Chongyi and Wang, Shiqi and Kwong, Sam},
  booktitle={Proceedings of the IEEE/CVF Conference on Computer Vision and Pattern Recognition},
  pages={1694--1704},
  year={2024}
}

@article{dino,
  title={Dinov2: Learning robust visual features without supervision},
  author={Oquab, Maxime and Darcet, Timoth{\'e}e and Moutakanni, Th{\'e}o and Vo, Huy and Szafraniec, Marc and Khalidov, Vasil and Fernandez, Pierre and Haziza, Daniel and Massa, Francisco and El-Nouby, Alaaeldin and others},
  journal={arXiv preprint arXiv:2304.07193},
  year={2023}
}

@article{chen2024sharegpt4video,
title={ShareGPT4Video: Improving Video Understanding and Generation with Better Captions},
author={Chen, Lin and Wei, Xilin and Li, Jinsong and Dong, Xiaoyi and Zhang, Pan and Zang, Yuhang and Chen, Zehui and Duan, Haodong and Lin, Bin and Tang, Zhenyu and Yuan, Li and Qiao, Yu and Lin, Dahua and Zhao, Feng and Wang, Jiaqi},
journal={arXiv preprint arXiv:2406.04325},
year={2024}
}

@inproceedings{sb,
  title={Swiftbrush: One-step text-to-image diffusion model with variational score distillation},
  author={Nguyen, Thuan Hoang and Tran, Anh},
  booktitle={Proceedings of the IEEE/CVF Conference on Computer Vision and Pattern Recognition},
  pages={7807--7816},
  year={2024}
}

@inproceedings{sbv2,
  title={Swiftbrush v2: Make your one-step diffusion model better than its teacher},
  author={Dao, Trung and Nguyen, Thuan Hoang and Le, Thanh and Vu, Duc and Nguyen, Khoi and Pham, Cuong and Tran, Anh},
  booktitle={European Conference on Computer Vision},
  pages={176--192},
  year={2024},
  organization={Springer}
}

@inproceedings{snoopi,
  title={Supercharged One-step Text-to-Image Diffusion Models with Negative Prompts},
  author={Nguyen, Viet and Nguyen, Anh and Dao, Trung and Nguyen, Khoi and Pham, Cuong and Tran, Toan and Tran, Anh},
  booktitle={Proceedings of the IEEE/CVF International Conference on Computer Vision},
  pages={18004--18013},
  year={2025}
}

@inproceedings{sc,
  title={Improved Training Technique for Shortcut Models},
  author={Nguyen, Anh and Van Nguyen, Viet and Vu, Duc and Dao, Trung Tuan and Tran, Chi and Tran, Toan and Tran, Anh Tuan},
  booktitle={The Thirty-ninth Annual Conference on Neural Information Processing Systems}
}

@inproceedings{efhq,
  title={Efhq: Multi-purpose extremepose-face-hq dataset},
  author={Dao, Trung Tuan and Vu, Duc Hong and Pham, Cuong and Tran, Anh},
  booktitle={Proceedings of the IEEE/CVF Conference on Computer Vision and Pattern Recognition},
  pages={22605--22615},
  year={2024}
}

@inproceedings{suma,
  title={SuMa: A Subspace Mapping Approach for Robust and Effective Concept Erasure in Text-to-Image Diffusion Models},
  author={Nguyen, Kien and Tran, Anh and Pham, Cuong},
  booktitle={Proceedings of the IEEE/CVF International Conference on Computer Vision},
  pages={19587--19596},
  year={2025}
}

@inproceedings{erase,
  title={Erasing concepts from diffusion models},
  author={Gandikota, Rohit and Materzynska, Joanna and Fiotto-Kaufman, Jaden and Bau, David},
  booktitle={Proceedings of the IEEE/CVF international conference on computer vision},
  pages={2426--2436},
  year={2023}
}

@article{nguyen2025cgce,
  title={CGCE: Classifier-Guided Concept Erasure in Generative Models},
  author={Nguyen, Viet and Patel, Vishal M},
  journal={arXiv preprint arXiv:2511.05865},
  year={2025}
}

@inproceedings{dreambooth,
  title={Dreambooth: Fine tuning text-to-image diffusion models for subject-driven generation},
  author={Ruiz, Nataniel and Li, Yuanzhen and Jampani, Varun and Pritch, Yael and Rubinstein, Michael and Aberman, Kfir},
  booktitle={Proceedings of the IEEE/CVF conference on computer vision and pattern recognition},
  pages={22500--22510},
  year={2023}
}

@article{textualinv,
  title={An image is worth one word: Personalizing text-to-image generation using textual inversion},
  author={Gal, Rinon and Alaluf, Yuval and Atzmon, Yuval and Patashnik, Or and Bermano, Amit H and Chechik, Gal and Cohen-Or, Daniel},
  journal={arXiv preprint arXiv:2208.01618},
  year={2022}
}

@misc{qwen,
      title={Qwen2.5 Technical Report}, 
      author={Qwen and : and An Yang and Baosong Yang and Beichen Zhang and Binyuan Hui and Bo Zheng and Bowen Yu and Chengyuan Li and Dayiheng Liu and Fei Huang and Haoran Wei and Huan Lin and Jian Yang and Jianhong Tu and Jianwei Zhang and Jianxin Yang and Jiaxi Yang and Jingren Zhou and Junyang Lin and Kai Dang and Keming Lu and Keqin Bao and Kexin Yang and Le Yu and Mei Li and Mingfeng Xue and Pei Zhang and Qin Zhu and Rui Men and Runji Lin and Tianhao Li and Tianyi Tang and Tingyu Xia and Xingzhang Ren and Xuancheng Ren and Yang Fan and Yang Su and Yichang Zhang and Yu Wan and Yuqiong Liu and Zeyu Cui and Zhenru Zhang and Zihan Qiu},
      year={2025},
      eprint={2412.15115},
      archivePrefix={arXiv},
      primaryClass={cs.CL},
      url={https://arxiv.org/abs/2412.15115}, 
}

@article{wan,
  title={Wan: Open and advanced large-scale video generative models},
  author={Wan, Team and Wang, Ang and Ai, Baole and Wen, Bin and Mao, Chaojie and Xie, Chen-Wei and Chen, Di and Yu, Feiwu and Zhao, Haiming and Yang, Jianxiao and others},
  journal={arXiv preprint arXiv:2503.20314},
  year={2025}
}
}

\clearpage
\setcounter{page}{1}
\maketitlesupplementary

\noindent\textbf{Overview:} We first introduce preliminaries on the CIELAB ($L^*a^*b^*$) color space in \cref{sec:lab} and the Discrete Cosine Transform (DCT) in \cref{sec:dct}. \cref{sec:implement} then outlines the experimental settings and parameters for all methods. \cref{sec:supprobust} details the hyperparameter settings for the purification and transformation techniques described in \cref{sec:robust}. In addition, \cref{sec:ablatedsp,sec:labbudget,sec:supptransfer,sec:loss,sec:ircloss,sec:promptvariety} provide ablation studies on the design of the perturbation optimization space, further transferability experiments, a component analysis of $\mathcal{L}_{Anti-I2V}$, and the method's effectiveness across diverse prompts. We additionally provide more details on the benchmark construction in \cref{sec:bench} and qualitative examples in \cref{sec:qualextra}.

\noindent\textbf{Note:} Ablation experiments in the supplementary material are conducted using CogVideoX-5B \cite{cogX} on a subset of 200 videos from CelebV-Text \cite{celebvtext}, where the first frame serves as the image condition and the provided caption serves as the prompt. For each image–prompt pair, we generate five samples, resulting in a total of \textbf{1,000 videos} for evaluation. Unless stated otherwise, all experiments follow this setup.

\section{CIELAB ($L^*a^*b^*$) color space}
\label{sec:lab}
The conversion from linear RGB (normalized to [0, 1]) to L\*a\*b\* is a two-stage process. First, a linear transformation to CIE XYZ space is performed using a predefined matrix $\mathbf{M}$: $\begin{bmatrix} X & Y & Z \end{bmatrix}^T = \mathbf{M} \begin{bmatrix} R & G & B \end{bmatrix}^T$. Subsequently, a non-linear transformation yields the L\*a\*b\* values.
\begin{equation}
\begin{aligned}
&X_n = 0.95047, \quad Y_n = 1.0, \quad Z_n = 1.08883,\\
&f(t) = \begin{cases} t^{1/3} & \text{if } t > (6/29)^3 \\ \frac{1}{3}(29/6)^2 t + 4/29 & \text{otherwise} \end{cases} ,\\
&L^* = 116 f(Y/Y_n) - 16, \\
&a^* = 500 [f(X/X_n) - f(Y/Y_n)], \\
&b^* = 200 [f(Y/Y_n) - f(Z/Z_n)].
\end{aligned}
\end{equation}

\section{Discrete Cosine Transform (DCT)}
\label{sec:dct}
For an RGB image \( x_0 \in \mathbb{R}^{3 \times h \times w} \), its frequency representation \( X_0 \) is given by:
\begin{equation}
    X_0(k, u, v) = c_u c_v \sum_{i=0}^{h-1} \sum_{j=0}^{w-1} x_0(k, i, j) \phi(i, u) \phi(j, v),
\end{equation}

where $k$ is the channel index, $u$ and $v$ are 2D coordinates in the frequency space, $c_u=\sqrt{1/h}$  if $u=0$ or $c_u=\sqrt{2/h}$  otherwise, $\phi(i,u)=\cos\left( \frac{\pi (0.5 + i)}{h} u \right)$, and $c_v$ and $\phi(j,v)$ have similar formulas as $c_u$ and $\phi(i,u)$, respectively. The inverse function, i.e., IDCT, to map from frequency domain to RGB domain is defined as:
\begin{equation}
    x_0(k, i, j) = \sum_{u=0}^{h-1} \sum_{v=0}^{w-1} c_u c_v X_0(k, u, v) \phi(i, u) \phi(j, v).
\end{equation}

\section{Implementation Details}
\label{sec:implement}
For all experiments, we fix the number of update iterations to $N = 200$ and use a perturbation budget of $\Delta_{RGB} = 16/255$. For our \textbf{Anti-I2V} method, we additionally set $\Delta_{Lab} = 16/255$. To improve efficiency and reduce memory usage, we use only the first four frames of each video as input to the VDMs. Following \cite{antiforgery}, we adopt the AdamW optimizer with a learning rate of $1\mathrm{e}{-2}$. Baseline methods are optimized using PGD \cite{pgd} with a step size of $1/255$. All experiments are run on a single NVIDIA A100 GPU 40GB.

\section{Robustness Settings}
\label{sec:supprobust}
For JPEG Compression, we compress each image at the compression rate of 40\%. For Gaussian blur, we set the kernel size to 7 and $\sigma=1.5$. For Gaussian noise, we set the noise scale to 0.05. For DIffPure \cite{diffpure}, we set the number of iterations to 100, with $\epsilon_{adv}=0.07$. For GrIDpure \cite{gridpure}, we also set the number of iterations to 100, with $\gamma=0.1$. All experiments use the same objective function, $\mathcal{L}_{Anti-I2V}$.

\section{Analysis of Perturbation Update Space}
\label{sec:ablatedsp}
To evaluate the Dual-Space Perturbation design (\cref{subsec:dsp}), we compare perturbations applied in RGB space, $L^*a^*b^*$ space, frequency space, and their combinations. For a fair comparison, all adversarial attacks use only the vanilla denoising loss as the optimization objective.

\begin{table}[H]
\centering
\footnotesize
\tabcolsep=0.05cm
\renewcommand\arraystretch{1}
\caption{\textbf{Quantitative results} of perturbation optimization spaces.}
\begin{tabular}{lcccc}
    \toprule
    {\textbf{Method}} &\textbf{ISM$\downarrow$} &\textbf{Q-A (F)$\downarrow$} &\textbf{Q-A (V)$\downarrow$} &\textbf{DINO-SIM$\downarrow$} \\
    \midrule
    RGB & \textbf{0.582} & 0.624 & 0.692 & 0.788 \\
    $L^*a^*b^*$ & 0.672 & 0.707 & 0.765 & 0.833 \\
    Frequency & 0.633 & 0.576 & 0.641 & 0.802 \\
    \midrule
    $\text{RGB + $L^*a^*b^*$}$ & 0.613 & 0.525 & 0.618 & 0.784 \\
    $\text{RGB + Frequency}$ & 0.587 & 0.567 & 0.645 & 0.796 \\
    \rowcolor{blue!12} $\textbf{$L^*a^*b^*$}$ + Frequency \textbf{(DSP)} & \textbf{0.582} & \textbf{0.521} & \textbf{0.610} & \textbf{0.781} \\
    \midrule
    $\text{RGB + DSP}$ & 0.654 & 0.566 & 0.639 & 0.805 \\
    \bottomrule	
\end{tabular}
\label{tab:DSP}
\end{table}

\noindent As shown in \cref{tab:DSP}, under identical objectives and settings, our \textbf{Dual-Space Perturbation (DSP)} yields stronger cloaking effects than traditional RGB-space perturbations. While perturbing only in the $L^*a^*b^*$ or frequency domains underperforms RGB, combining these domains substantially improves protection, as reflected by the drops in Q-Align (V) and Q-Align (F). Adding RGB to DSP, however, degrades performance because the fixed perturbation budget must be split across more spaces, reducing the impact of each. Although RGB performs comparably on ISM, it falls short in other overall quality and aesthetics metrics. DSP is preferred for its stronger robustness to purification, as elaborated in \cref{sec:robust}

\section{Perturbation budget for $L^*a^*b^*$ color space}
\label{sec:labbudget}
\begin{table}[t]
\centering
\footnotesize
\tabcolsep=0.20cm
\caption{\textbf{Analysis of perturbation budget for $L^*a^*b^*$ space.}}
\renewcommand\arraystretch{1}
\begin{tabular}{lccccc}
    \toprule
    {\textbf{Budget}} & \textbf{ISM$\downarrow$} & \textbf{C-FIQA $\downarrow$} & \textbf{Q-A (F)$\downarrow$} &\textbf{Q-A (V)$\downarrow$} &\textbf{DINO$\downarrow$} \\
    \midrule
    \textbf{$8/255$} & 0.469 & 0.456 & 0.485 & 0.567 & 0.775 \\
    \rowcolor{blue!12}\textbf{$16/255$} & \textbf{0.462} & \textbf{0.448} & \textbf{0.481} & \textbf{0.562} & \textbf{0.760} \\
    \textbf{$32/255$} & 0.473 & 0.468 & 0.491 & 0.588 & 0.775 \\
    \bottomrule	
\end{tabular}
\label{tab:lab}
\end{table}
\cref{tab:lab} shows the performance of our method under different perturbation budgets in the $L^*a^*b^*$ color space. With all other parameters and loss components fixed and using the same objective function, $\mathcal{L}_{Anti-I2V}$, we evaluate three levels of $\Delta_{Lab}$: $8/255$, $16/255$, and $32/255$. Increasing the perturbation budget does not always enhance protection. The best balance between identity concealment and video quality is achieved at $\Delta_{Lab} = 16/255$, which achieves the lowest DINO-SIM while maintaining the lowest ISM and Q-Align scores. This suggests that $\Delta_{Lab}$ can be flexibly selected based on the desired protection and objective.

\section{Transferability}
\label{sec:supptransfer}
Following the protocol in \cref{sec:transfer}, we additionally evaluate transferability on the recent Wan2.2-TI2V-5B~\cite{wan}, as shown in \cref{tab:cogwan}. Consistent with the other transfer settings, Anti-I2V clearly surpasses all baselines on Wan2.2, further demonstrating strong generalization to up-to-date models.

\begin{table}[b]
    \centering
    \small
    \setlength{\tabcolsep}{3.5pt} 
    \caption{\textbf{Quantitative comparison of transferbility} from CogVideoX-5B to Wan2.2-TI2V-5B.}
    \label{tab:cogwan}
        \begin{tabular}{lccccc}  
        \toprule
        & \multicolumn{5}{c}{\textbf{CogVideoX-5B - Wan2.2-TI2V-5B}} \\
        \cmidrule(lr){2-6}
        Method & ISM $\downarrow$ & C-FIQA $\downarrow$ & Q-A(F) $\downarrow$ & Q-A(V) $\downarrow$ & DINO $\downarrow$ \\
        \midrule
        Clean     & 0.672 & 0.517 & 0.841 & 0.899 & 0.815 \\
        SDS+      & 0.538 & 0.478 & 0.512 & \textbf{0.622} & \textbf{0.741} \\
        SDS-      & 0.608 & 0.504 & 0.573 & 0.667 & 0.766 \\
        AdvDM     & 0.544 & 0.456 & 0.528 & 0.624 & \underline{0.743} \\
        MIST      & 0.635 & 0.476 & 0.574 & 0.678 & 0.790 \\
        VGMShield & 0.628 & 0.499 & 0.530 & 0.624 & 0.778 \\
        \rowcolor{blue!12}\textbf{Anti-I2V} 
                  & \textbf{0.439} & \textbf{0.450} & \textbf{0.502} & \underline{0.623} & \underline{0.743} \\
        \bottomrule
        \end{tabular}
\end{table}

\section{Loss Components}
\label{sec:loss}
\cref{tab:loss} highlights the effectiveness of applying IRC and IRA compared to the vanilla denoising loss. Specifically, both IRC and IRA, when individually combined with the vanilla denoising loss, lead to a decline across all metrics, particularly in identity-related features. This indicates that these components disrupt the information flow within the denoising modules, causing the reference image features to diverge. Furthermore, the results suggest that high-level semantic features (e.g., human identity) are significantly affected. Combining both losses further reduces ISM and DINO-SIM, implying that IRC and IRA complement each other.

\begin{table}[t]
\centering
\renewcommand{\arraystretch}{1.1}
\caption{\textbf{Ablation Study of Loss Components}: 
Comparison of different loss components on performance metrics. (U) denotes untargeted attack, while (T) denotes targeted attack.}
\resizebox{\linewidth}{!}{
\begin{tabular}{lcccc}
\toprule
\textbf{Loss Type} & ISM $\downarrow$ & Q-A (F) $\downarrow$ & Q-A (V) $\downarrow$ & DINO $\downarrow$ \\
\midrule
\textbf{[A1]}: Denoising Loss & 0.582 & 0.521 & 0.610 & 0.781 \\
\midrule
\textbf{[A2]}: \textbf{[A1]} + IRC & 0.535 & 0.518 & 0.607 & 0.776 \\
\midrule
\textbf{[A3]}: \textbf{[A1]} + IRA-VAE (U) & 0.514 & 0.507 & 0.583 & 0.774 \\
\textbf{[A4]}: \textbf{[A1]} + IRA-VAE (T) & 0.507 & 0.503 & 0.580 & 0.771 \\
\midrule
\textbf{[A5]}: \textbf{[A1]} + IRA-Denoiser (U) & 0.507 & 0.486 & 0.572 & 0.775 \\
\textbf{[A6]}: \textbf{[A1]} + IRA-Denoiser (T) & 0.493 & 0.484 & 0.575 & 0.772 \\
\midrule
\textbf{[A7]}: \textbf{[A4]} + \textbf{[A6]} & 0.484 & 0.483 & 0.567 &  0.768 \\
\midrule
\textbf{[A8]}: \textbf{[A2]} + \textbf{[A4]} + \textbf{[A6]} & 0.476 & 0.481 &  0.567 & 0.764 \\
\midrule
\textbf{[A9]}: \textbf{[A8]} + Auxiliary Loss & \textbf{0.462} & \textbf{0.481} & \textbf{0.562} & \textbf{0.760} \\
\bottomrule
\end{tabular}}
\label{tab:loss}
\end{table}

\section{IRC Layer Selection}
\label{sec:ircloss}
\cref{tab:irclayer} highlights the effectiveness of applying IRC under different layer configurations. We compare the optimal layer selection identified in \cref{subset:IRC} with a simplified variant that applies the IRC loss only on the last three layers. We further investigate more configurations by applying IRC to the last one, two, and four layers. The results show virtually identical performance, indicating that the simplified setting of IRC loss provides comparable protection without any need of complicated layer selection.

\begin{table}[H]
\centering
\scriptsize
\renewcommand{\arraystretch}{1.15}
\caption{\textbf{Ablation on IRC Layer Selection}. 
We compare applying IRC to different subsets of layers. 
\textbf{Full} applies IRC to all layers after the 27\textsuperscript{th} layer. 
\textbf{Last-$k$} applies IRC only to the final $k$ layers. 
\textbf{Bold} indicates the best performance, and \underline{underline} indicates the second best.}
\resizebox{\linewidth}{!}{
\begin{tabular}{lcccc}
\hline
\textbf{Setting} & ISM $\downarrow$ & Q-A (F) $\downarrow$ & Q-A (V) $\downarrow$ & DINO $\downarrow$ \\ 
\hline

Full (27+) 
& \textbf{0.458} & \textbf{0.479} & \textbf{0.560} & 0.762 \\

Last-3 
& 0.462 & \underline{0.481} & \underline{0.562} & \textbf{0.760} \\

Last-1 
& \underline{0.460} & 0.486 & 0.565 & 0.762 \\

Last-2 
& \underline{0.460} & 0.487 & 0.568 & 0.767 \\

Last-4 
& 0.470 & 0.489 & 0.570 & 0.764 \\
\hline
\end{tabular}}
\label{tab:irclayer}
\end{table}

\section{Evaluation on Different Prompts}
\label{sec:promptvariety}
We evaluate whether our method can generate effective attacks independent of the provided prompts. For each image ID in our evaluation subset, we use \cite{chen2024sharegpt4video} to generate three distinct text prompts. Captions are obtained using the query: "Return me three different text prompts for video generation based on this image. The prompts should focus on the human subject, their appearance, and their actions." For each image–prompt pair, we generate five samples, resulting in a total of 3000 videos for evaluation. We use CogVideoX \cite{cogX} and DynamiCrafter \cite{dynamic} as representative models for DiT-based and UNet-based architectures, respectively. \cref{tab:prompt} demonstrates that despite the difference in provided prompts, our method significantly successfully degrades both identity features and video quality.Our method consistently achieves strong protection across different prompts, as evidenced by lower ISM, CLIB-FIQA, Q-Align, and DINO-SIM scores. Moreover, our method maintains its effects on DynamiCrafter \cite{dynamic}, similar to experiments in \cref{sec:exp}. This suggests that our approach is robust to prompt variations, effectively disrupting video generation regardless of the textual descriptions used.

\begin{table}[H]
    \caption{\textbf{Quantitative comparisons of protections with different set of prompts.} $\downarrow$ indicates that a lower value of the metric signifies poorer video quality and thus better protection.}
    \label{tab:prompt}
    \footnotesize
    \tabcolsep=0.30cm
    \begin{tabular}{l l c c}
        \toprule
        {\textbf{Method}} & \textbf{Metric} $\downarrow$ & \textbf{Clean} & \cellcolor{blue!15} {\textbf{Anti-I2V}} \\
        \midrule
        \multirow{5}{*}{\textbf{CogVideoX} ~\cite{cogX}} &  
        ISM & 0.646 & \textbf{0.407} \\
        & C-FIQA & 0.519 & \textbf{0.462} \\
        & Q-A (Frame) & 0.771 & \textbf{0.474} \\
        & Q-A (Video) &  0.825 & \textbf{0.553} \\
        & DINO & 0.869 & \textbf{0.776} \\
        \midrule
        \multirow{5}{*}{\textbf{DynamiCrafter~\cite{dynamic}}} & 
        ISM & 0.558 & \textbf{0.208} \\
        & C-FIQA & 0.521 & \textbf{0.367} \\
        & Q-A (Frame) & 0.883 & \textbf{0.084} \\
        & Q-A (Video) & 0.912 & \textbf{0.104} \\
        & DINO & 0.875 & \textbf{0.234} \\
        \bottomrule	
    \end{tabular}
\end{table}

\section{Perturbation Visibility}
We conduct experiments to evaluate the imperceptibility of each method, using SSIM and PSNR as our primary metrics. Notably, perturbations in the Lab color space prioritize perceptually meaningful changes, which do not fully align with how PSNR and SSIM measure image similarity. These metrics emphasize pixel-wise differences and structural consistency in the RGB space, making them less reflective of human visual perception. As a result, while perturbations in the Lab space subtly alter color information in a way that is less noticeable to humans, they can still lead to lower PSNR and SSIM scores due to significant pixel-level differences. Nevertheless, as shown in \cref{tab:visible}, our method remains competitive across all metrics despite its lower SSIM and PSNR values.

\begin{table}[H]
\centering
\scriptsize
\begin{tabular}{lcccc}
\toprule
\textbf{Method} & \textbf{SSIM}$\uparrow$ & \textbf{LPIPS}$\downarrow$ & \textbf{PSNR}$\uparrow$ \\
\midrule
SDS (+) & 0.84 & 0.206 & 32.5 \\
SDS (-) & \textbf{0.86} & 0.192 & \textbf{33.7} \\
AdvDM & 0.84 & 0.205 & 32.6 \\
MIST  & 0.82 & 0.271 & 31.4 \\
VGMShield & 0.84 & \textbf{0.191} & 33.2 \\
Ours & 0.80 & 0.200 & 32.2 \\
\bottomrule
\end{tabular}
\caption{\textbf{Ablation Study of Perturbation Visibility}: 
Similarity metrics between perturbed images and their original images of different methods.}
\label{tab:visible}
\end{table}

\section{Benchmark Construction}
\label{sec:bench}
To obtain reference videos for the inputs, we use~\cite{chen2024sharegpt4video} to generate captions describing the adversarial examples. Specifically, we query the model with the prompt: "Return an extremely detailed prompt ONLY describing the person in this image (including appearance, emotion, and action)." For CelebV-Text~\cite{celebvtext}, we first crawl videos with unique identities, using Qwen \cite{qwen} to obtain person-centric descriptions, from which we synthesize the first-frame images using FLUX~\cite{flux}. We then pair each image with the corresponding video prompt described above, crop the image to satisfy the model input requirements, and generate five samples for each image--prompt pair.

\section{More qualitative results}
\label{sec:qualextra}
\begin{figure*}[!t]
    \centering
    \includegraphics[width=\linewidth]{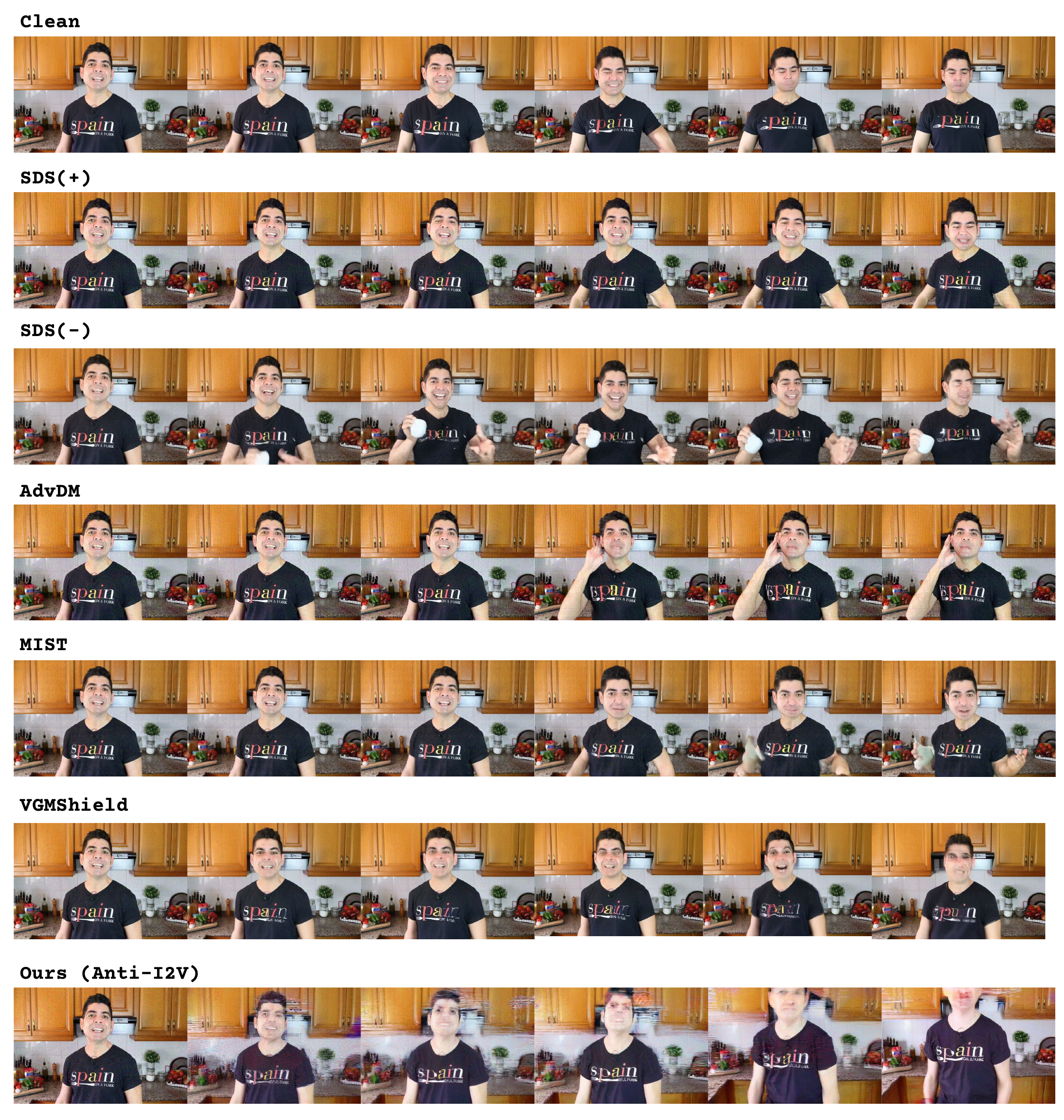} 
    \caption{Qualitative comparison of adversarial attack methods against against CogVideoX \cite{cogX}. The first column shows the reference frame. The remaining columns present the generated outputs from models.}
    \label{fig:qualicog1}
\end{figure*}

\begin{figure*}[!t]
    \centering
    \includegraphics[width=\linewidth]{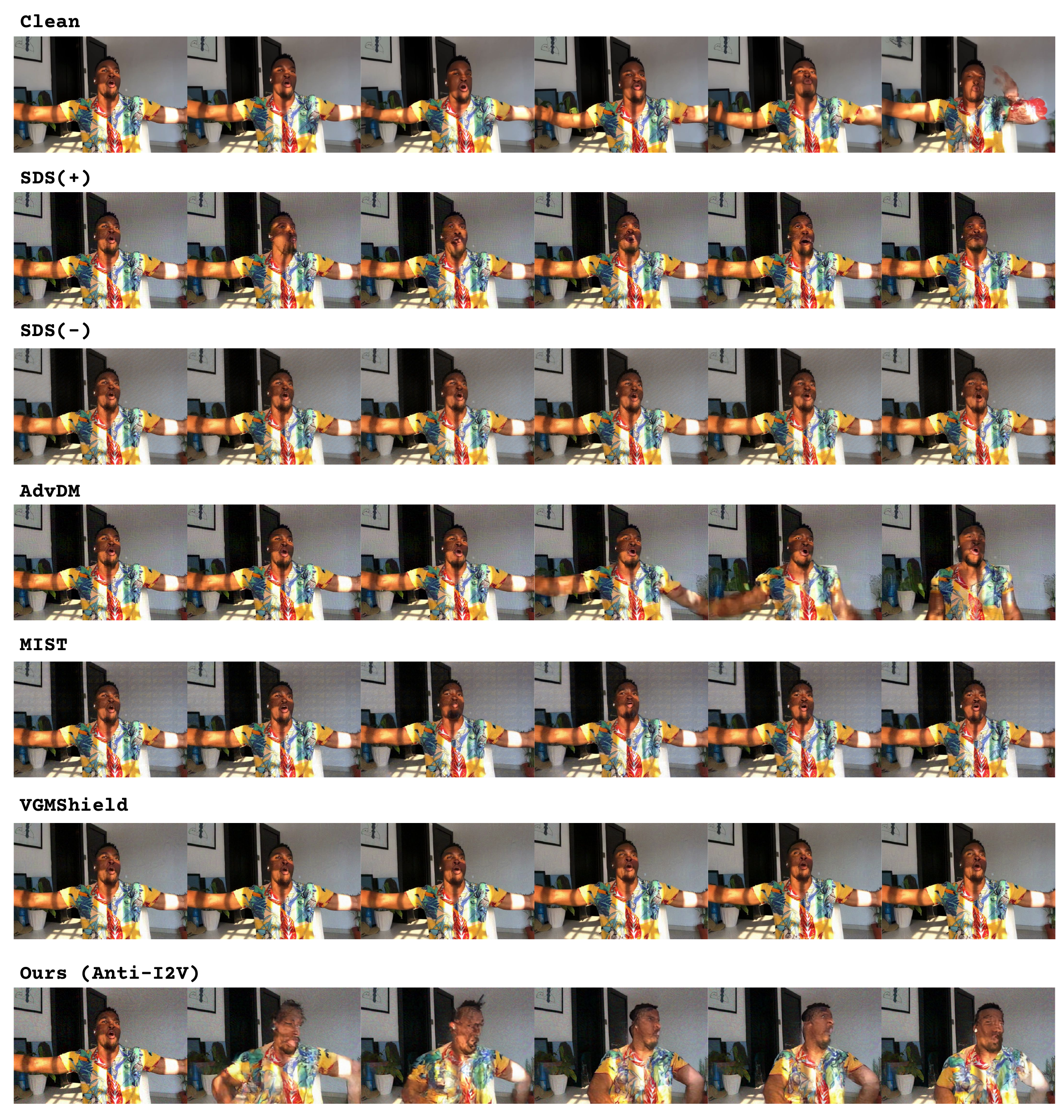} 
    \caption{Qualitative comparison of adversarial attack methods against against CogVideoX \cite{cogX}. The first column shows the reference frame. The remaining columns present the generated outputs from models.}
    \label{fig:qualicog2}
\end{figure*}

\begin{figure*}[!t]
    \centering
    \includegraphics[width=\linewidth]{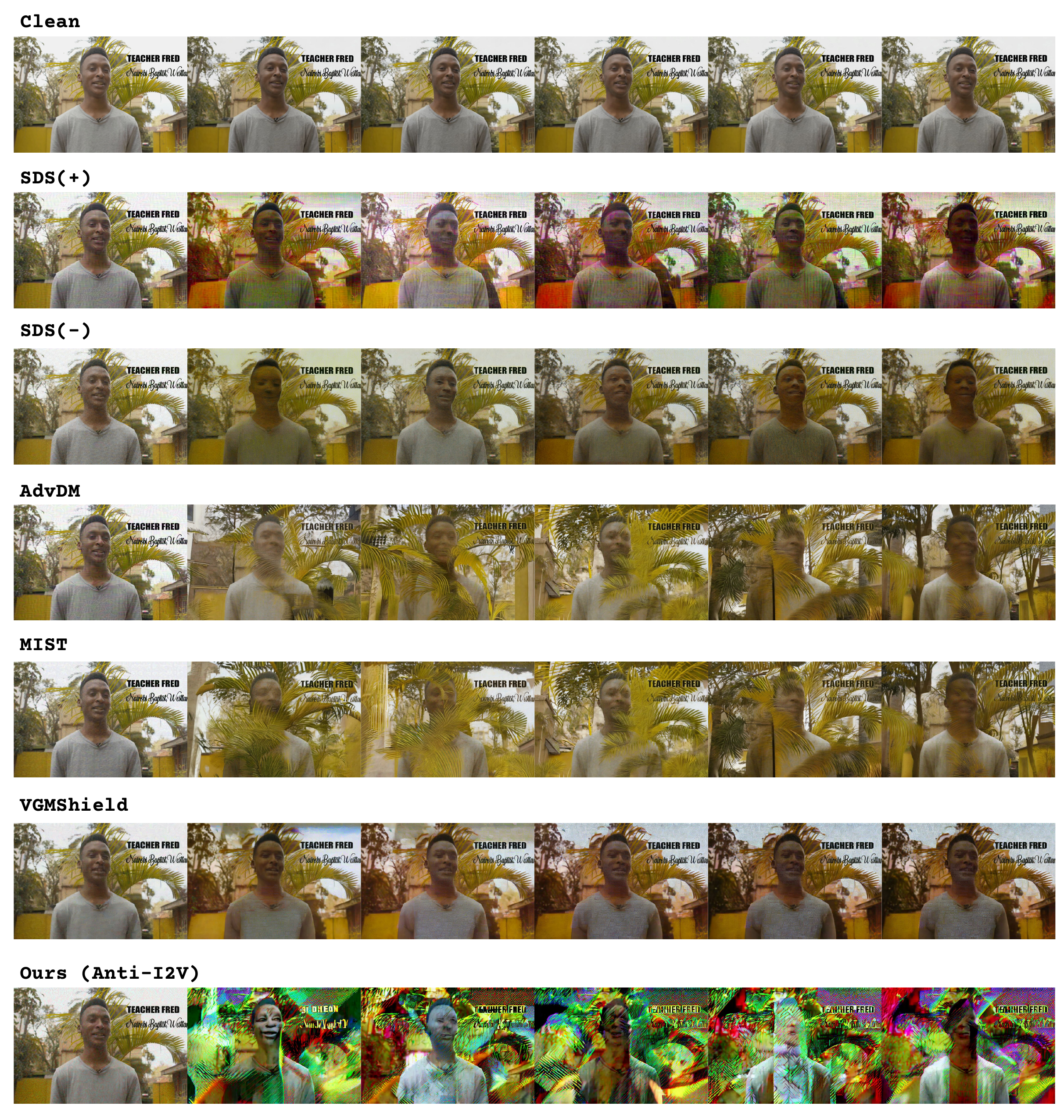} 
    \caption{Qualitative comparison of adversarial attack methods against against DynamiCrafter \cite{dynamic}. The first column shows the reference frame. The remaining columns present the generated outputs from models.}
    \label{fig:qualidyn1}
\end{figure*}

\begin{figure*}[!t]
    \centering
    \includegraphics[width=\linewidth]{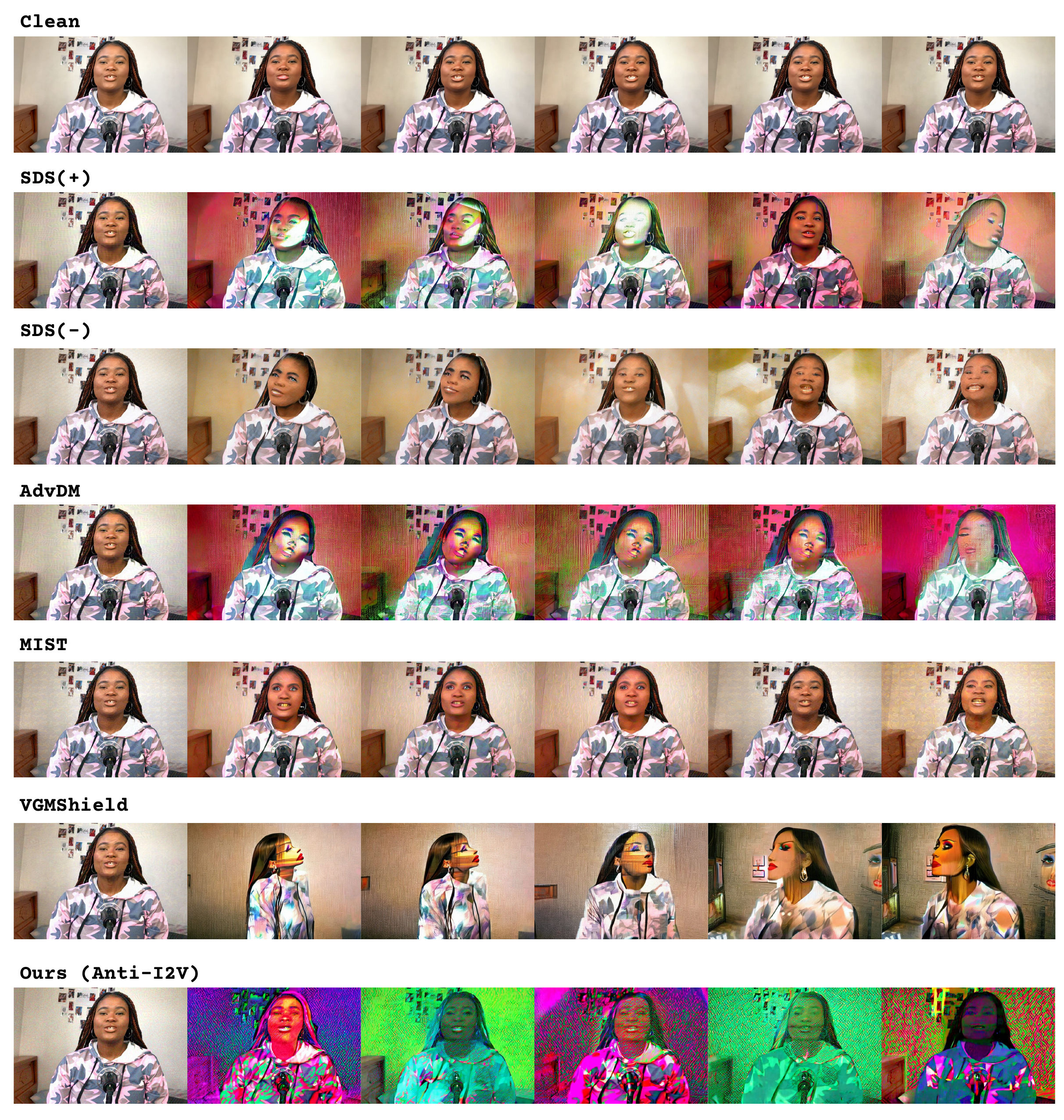} 
    \caption{Qualitative comparison of adversarial attack methods against against DynamiCrafter \cite{dynamic}. The first column shows the reference frame. The remaining columns present the generated outputs from models.}
    \label{fig:qualidyn2}
\end{figure*}

We provide more qualitative examples to compare our method Anti-I2V against other baselines with different diffusion models. Please refer to \cref{fig:qualicog1,fig:qualicog2,fig:qualidyn1,fig:qualidyn2}.

\end{document}